\documentclass[11pt]{article}

\usepackage[letterpaper, margin=0.98in, top=1in, bottom=1in]{geometry}

\usepackage{microtype}
\usepackage{graphicx}
\usepackage{subfigure}
\usepackage{booktabs} 
\usepackage{hyperref}

\newcommand{\removed}[1]{}
\usepackage[textsize=tiny]{todonotes}
\newcommand{\natinote}[1]{\todo[color=blue!20]{{#1}}}

\newcommand{\marko}[1]{{\color{olive} [Marko: #1]}}

\newcommand{\kaifeng}[1]{{\color{orange} [KL:#1]}}

\usepackage{amsmath}
\usepackage{amssymb}
\usepackage{mathtools}
\usepackage{amsthm}
\usepackage[capitalize,noabbrev]{cleveref}
\usepackage{bbm}
\usepackage{natbib}
\usepackage{comment}

\usepackage{aliascnt}
\usepackage[capitalize,noabbrev]{cleveref} 

\theoremstyle{plain}
\newtheorem{theorem}{Theorem}[section]

\newaliascnt{lemma}{theorem}
\newtheorem{lemma}[lemma]{Lemma}
\aliascntresetthe{lemma}
\crefname{lemma}{lemma}{lemmas}
\Crefname{lemma}{Lemma}{Lemmas}

\newaliascnt{proposition}{theorem}
\newtheorem{proposition}[proposition]{Proposition}
\aliascntresetthe{proposition}
\crefname{proposition}{proposition}{propositions}
\Crefname{proposition}{Proposition}{Propositions}

\newaliascnt{corollary}{theorem}
\newtheorem{corollary}[corollary]{Corollary}
\aliascntresetthe{corollary}
\crefname{corollary}{corollary}{corollaries}
\Crefname{corollary}{Corollary}{Corollaries}

\theoremstyle{definition}

\newaliascnt{definition}{theorem}
\newtheorem{definition}[definition]{Definition}
\aliascntresetthe{definition}
\crefname{definition}{definition}{definitions}
\Crefname{definition}{Definition}{Definitions}

\newaliascnt{assumption}{theorem}

\aliascntresetthe{assumption}
\crefname{assumption}{Condition}{Conditions}
\Crefname{assumption}{Condition}{Conditions}

\newaliascnt{condition}{theorem}
\newtheorem{condition}[condition]{Condition}
\aliascntresetthe{condition}
\crefname{condition}{Condition}{Conditions}
\Crefname{condition}{Condition}{Conditions}

\newaliascnt{example}{theorem}

\aliascntresetthe{example}
\crefname{example}{example}{examples}
\Crefname{example}{Example}{Examples}

\newaliascnt{notation}{theorem}

\aliascntresetthe{notation}
\crefname{notation}{notation}{notations}
\Crefname{notation}{Notation}{Notations}

\newaliascnt{model}{theorem}
\newtheorem{model}[model]{Model}
\aliascntresetthe{model}
\crefname{model}{model}{models}
\Crefname{model}{Model}{Models}

\theoremstyle{remark}
\newaliascnt{remark}{theorem}

\aliascntresetthe{remark}
\crefname{remark}{remark}{remarks}
\Crefname{remark}{Remark}{Remarks}

\usepackage{enumitem}
\setlist[enumerate,1]{leftmargin=0.6cm}
\setlist[itemize,1]{leftmargin=0.4cm}
\usepackage{amsmath,amsfonts,bm}

\def\1{\bm{1}}

\def\vzero{{\bm{0}}}

\def\va{{\bm{a}}}

\makeatletter
\newcommand{\ve}{\@ifnextchar\bgroup{\velong}{{\bm{e}}}}
\newcommand{\velong}[1]{{\bm{#1}}}
\makeatother

\def\vn{{\bm{n}}}

\def\vp{{\bm{p}}}
\def\vq{{\bm{q}}}
\def\vr{{\bm{r}}}
\def\vs{{\bm{s}}}

\def\vw{{\bm{w}}}
\def\vx{{\bm{x}}}

\def\vz{{\bm{z}}}

\def\mA{{\bm{A}}}

\def\mT{{\bm{T}}}

\DeclareMathAlphabet{\mathsfit}{\encodingdefault}{\sfdefault}{m}{sl}
\SetMathAlphabet{\mathsfit}{bold}{\encodingdefault}{\sfdefault}{bx}{n}

\def\gA{{\mathcal{A}}}

\def\gD{{\mathcal{D}}}

\def\gG{{\mathcal{G}}}

\newcommand{\E}{\mathbb{E}}
\newcommand{\R}{\mathbb{R}}

\DeclareMathOperator*{\argmin}{arg\,min}

\newcommand{\cD}{\mathcal{D}}

\newcommand{\cJ}{\mathcal{J}}

\newcommand{\cL}{\mathcal{L}}

\newcommand{\Z}{\mathbb{Z}}

\newcommand{\onec}[1]{\mathbbm{1}_{\{#1\}}}

\newcommand{\abs}[1]{\lvert #1 \rvert}
\newcommand{\labs}[1]{\left\lvert #1 \right\rvert}

\newcommand{\inne}[2]{\langle{#1}, {#2}\rangle}

\newcommand{\Dtrain}{D_{\mathrm{train}}}
\newcommand{\Dtest}{D_{\mathrm{test}}}

\newcommand{\Lsame}{L^{\mathrm{same}}}
\newcommand{\hLsame}{\hat{L}^{\mathrm{same}}}
\newcommand{\Lshift}{L_N}
\newcommand{\msame}{N^{\mathrm{same}}}
\newcommand{\mshift}{N^{\mathrm{shifted}}}
\newcommand{\mratio}{N^{\mathrm{ratio}}}

\makeatletter
\renewcommand{\paragraph}{%
  \@startsection{paragraph}{4}%
  {\z@}{0ex}{-1em}%
  {\normalfont\normalsize\bfseries}%
}
\makeatother

\Crefname{assumption}{Condition}{Conditions}

\title{Shift is Good: Mismatched Data Mixing \\ Improves Test Performance}
\makeatletter
\renewcommand{\maketitle}{%
  \begin{center}
    {\LARGE\bfseries \@title \par}
  \end{center}
  \vspace{0.8em}
  {\raggedright\large \@author \par}
}
\makeatother

\author{%
\textbf{Marko Medvedev}\textsuperscript{*}\textsuperscript{$\dagger$}\\
University of Chicago\\[0.75em]
\textbf{Kaifeng Lyu}\textsuperscript{*}\\
Tsinghua University\\[0.75em]
\textbf{Zhiyuan Li}\\
\textbf{Nathan Srebro}\\
Toyota Technological Institute at Chicago
}

\date{}

\begin{document}
\maketitle

{\begingroup
  \renewcommand{\thefootnote}{\fnsymbol{footnote}}
  \NoHyper\footnotetext[1]{Equal Contribution}\endNoHyper
  \NoHyper\footnotetext[2]{Email: \texttt{\{last name\}@uchicago.edu}}\endNoHyper
\endgroup}
\setcounter{footnote}{0}

\begin{abstract}
    We consider training and testing on mixture distributions with different training and test proportions.  We show that in many settings, and in some sense generically, distribution shift can be beneficial, and test performance can improve due to mismatched training proportions, even if the components are unrelated and with no transfer between components.  In a variety of scenarios, we identify the optimal training proportions and the extent to which such distribution shift can be beneficial.  We show how the same analysis applies also to a compositional setting with differing distribution of component ``skills'' at training and test.
\end{abstract}

\section{Introduction}\label{sec:introduction}



Imagine that you are taking a high-stakes exam next week. The exam will be 90\% on European history and 10\% on Chinese history.  Both topics are equally familiar to you and equally difficult, and additional study will help you with each topic similarly.  You have unlimited access to study material and practice questions for both.  How should you spend your limited studying budget?  Should your training match your test distribution, studying 90\% European and 10\% Chinese?  Or would you benefit from a distribution shift?  Studying more Chinese history?  Less?  Only European history? \emph{We encourage the reader to pause and make an intuitive guess.}

The answer depends on the specific learning curve for improvement in test performance within a topic as a function of the number of training examples from that topic.  But at least for a generic $1/n$ scaling (as obtained from e.g., both learning VC classes and in parametric regression), the answer, as we will see in \Cref{sec:orthogonalpowerlaw}, is that you would benefit from a distribution shift, and should study 75\% European History and 25\% Chinese history---this would reduce your test error by 20\% over the 90\%/10\% non-shifted training.

\begin{figure*}[h]
    \vspace{-0.1in}
    \includegraphics[width=0.4\textwidth]{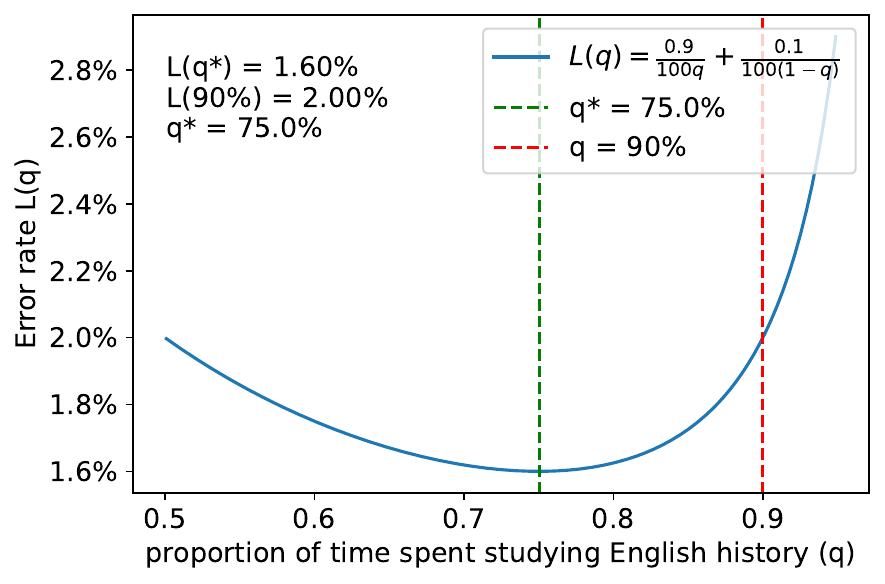}
    \includegraphics[width=0.4\textwidth]{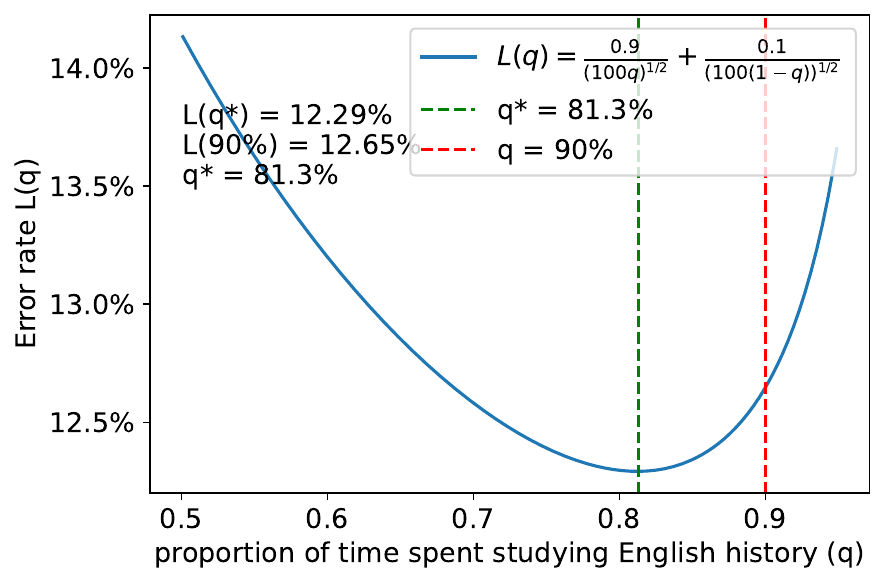}
    \centering
    \vspace{-0.15in}
    \caption{
    \small    
    We plot the error rate for a hypothetical scenario modelling the high stakes exam described in \Cref{sec:introduction}. We model the error rate on each of the test portions as being proportional to $\propto \frac{1}{n_i^{\alpha}}$, where $n_i$ represents the studying budget spent on that portion of the exam, so $i=1$ corresponds to European History and $i=2$ to the Chinese History and set $n_1+n_2=N$ to be the total studying budget, with $N=100$ hours. The exponent $\alpha$ is $\alpha=1$ on the left plot and $\alpha=2$ on the right plot. In both cases, we consider $n_1=qN$ and $n_2=(1-q)N$, where $q$ is the proportion of time spent studying for the European History portion of the exam. This way, the error rate on the exam can be written as a function of $q$ as $L(q) = 0.9\frac{1}{(100q)^{\alpha}}+0.1\frac{1}{(100q)^{\alpha}}$. We can see on both plots that shifting away from the testing proportion (red line, i.e. $q=90\%$) can lead to a better error rate with the optimal test proportion (green line, i.e., $q^*$ whose values are displayed accordingly). See also~\Cref{cor:main:generalpower}.}\label{fig:test-taking} 

    \vspace{-0.1in}
\end{figure*}

We just saw an example of what we term \textbf{Positive Distribution Shift}: Even if we have unlimited data from the target test distribution $\Dtest$, training on a shifted distribution $\Dtrain\neq\Dtest$ can actually {\em improve} test performance.  This contrasts the typical study of {\em distribution shift}, i.e.,~training on one distribution but then applying the predictor, or testing, on another. In that line of work, an implicit baseline would be to train on the test distribution, and any deviation to the case of $\Dtrain \ne \Dtest$ is viewed as a compromise.
This deviation may occur because we do not know or cannot directly access the true $\Dtest$, because it is too expensive to sample from $\Dtest$, or because we only have a limited number of samples and want to supplement them with additional data from related distributions.  
In the standard view, distribution shift is often posed as ``how much worse do things get if we train on $\Dtrain\neq\Dtest$?'' A typical answer is of the form: ``if $\Dtrain$ is close or related enough to $\Dtest$, then the performance is not much worse.''

        In this paper, we investigate one of several ways in which distribution shift can be \emph{positive}. We focus on the case when the test distribution is given as a mixture of $K$ components (or tasks), with known mixing proportions $\{ p_k \}_{k=1}^{K}$, and consider training distributions which are mixtures over the same components but with different mixing proportions $\{q_k\}_{k=1}^{K}$. We study how positive distribution shift can happen \emph{even though the tasks are independent} and there is no transfer, demonstrating that the improvement can arise purely from mixture effects. We systematically demonstrate the benefit of such distribution shift in terms of improved sample complexity when training with mismatched mixing proportions relative to the test distribution.  In fact, in \Cref{sec:general-case} we argue that positive distribution shift is the norm, rather then the exception, and almost always happens.
We can either think of our results as providing guidance when we can actively control mixing between different known components, or as helping us understand how and why a mismatched training distribution can actually be beneficial. 

In \Cref{sec:composition} we go beyond a mixture setting, and consider a compositional problem, where each instance involves composing multiple 'skills', with different skill frequencies.  E.g., solving mathematical problems with multiple simple steps, each of which is a 'skill'.  Should the training data have the same skill frequencies or different skill frequencies?  This problem, which we make concrete as a stylized LLM training problem, was a significant motivator for this research.  We show that when training on multiple skills with Chain-of-Thought training, although the setting is different, the effect of changing the skill distribution is related to that of the mixture setting, and thus our mixture setting analysis provides guidance here.  We show empirically the benefit of the predicted positive distribution shift in learning this stylized reasoning task.

In Section \ref{sec:transfer}, we depart from components for which learning is independent and consider a setting with {\em transfer} between the different components. For the most commonly studied transfer learning learning curves, we show that again, even though the error behaviour is different, the effect of mixture proportion mismatch is the same as for the independent learning setting we study, and so the results are applicable also here.

In this paper, we focus on how positive distribution shift can arise purely due to mixture effects, and not because of ``transfer'' between components or components being more or less informative or useful.  Positive distribution shift can certainly arrise also for other reason:  E.g., it might be better to train on cleaner or less noisy data, or more generally leveraging transfer patterns between tasks or components~\cite{albalak2023efficientonlinedatamixing, liu2025regmix, jiang2025adaptive, shukor2025scalinglawmixing}.  An even stronger benefit might be computational, where changes in the training distribution provide structure that is easier to exploit computationally, as is hinted by e.g.~\cite{abbe2023provable,wang2025learningcompositionalfunctionstransformers}. All of these effects can be compounded with the mixture effect we study here.  Indeed, several recent empirical papers looked at optimizing training mixture proportions in order to obtain good performance on a mixutre distribution such as Pile, showing empirically that the optimal training proportions differ from the test proportions \cite{xie2023doremioptimizingdatamixtures, ye2025mixing, albalak2023efficientonlinedatamixing,  jiang2025adaptive, shukor2025scalinglawmixing}.  As also indicated by the empirical scaling laws uncovered, this is due to a large part due to complex non-symetric transfer paterns between the different components, which could give rise to arbitrary Positive Distribution Shift paterns.   But in this paper, we focus on understanding and mathematically charactarizing the direct effect of changing the mixture proportions, both as an important effect in their own right, and to better disentangle them from other effects when understanding Positive Distribution Shift forces in more complex problems involving also transfer and computational aspects.  The papers mentioned here, as well as others \cite{gonzalez2015mismatched, hoffmann2022computeoptimal, sorscher2022beyond, xie2023data, gu2025data}, also emphasize the prevelance of Positive Distribution Shift in practice and how such ``data set selection'' is an important part of contemporary machine learning---it would thus benefit us to better understand and charactarize how and why it can happen and obtain a framework and language for discussing it.

\section{Setup}\label{sec:setup}

\paragraph{Learning Setup and Loss} Let $\ell(h, \vz)$ be the loss function that describes how well a model $h$ performs on an instance $\vz \in \mathcal{Z}$. For example, in supervised learning, $\vz$ can be an input-output pair $(\vx, y)$, and $\ell(h, \vz)$ can be the prediction error of $h(\vx)$ when $y$ is the ground truth.  Or, in next-word prediction, $\vz$ can be a document and $\ell(h, \vz)$ can be the average cross-entropy loss incurred when $h$ is used to predict each of the next tokens in the document. In any case, given a test distribution $\Dtest$ over $\vz$, we evaluate the model through the \emph{test loss} $\cL_{\Dtest}(h):=\E_{\vz \sim \Dtest}[\ell(h, \vz)]$.

\paragraph{Test Distribution.}
We consider test distributions that can be written as a mixture of $K$ components $\gD_1, \dots, \gD_K$. 
A mixture $\cD_\vr=\sum_k r_k \gD_k$ is determined by mixing proportions $\vr = (r_1,\dots,r_K) \in \Delta_K$, where $\Delta_K := \{ \vr \in \R^K : \vr \ge 0,\; \sum_{k=1}^{K} r_k = 1 \}$ denotes the probability simplex.
In the rest of the paper, we write $\vp$ for the mixing proportions of the test distribution, i.e.,~$\Dtest=\gD_\vp$, so the test loss is $\cL_{\gD_\vp}(h)=\cL_\vp(h)$,  where here and elsewhere we use the subscript $\vp$ on the loss to denote the mixture $\gD_\vp$.


\paragraph{Learning Algorithm.}
We consider an abstract ``learning algorithm'' $\gA$ that, given training data (or sequence of training examples) $S\in\mathcal{Z}^N$ of size $N$, produces a model $\gA(S)$. The performance of the model is evaluated with test loss $\cL_{\gD_\vp}(\gA(S))$.

\paragraph{Training Distribution.} We consider training on $N$ i.i.d.~samples $S\sim\cD_\vq^N$ from a mixture $\cD_\vq$ consisting of the same $K$ components as the test distribution, but with potentially different mixing proportions $\vq\in\Delta_K$.   For training mixing proportions $\vq$, we denote $L_N(\vp,\vq)=\E_{S \sim \cD_{\vq}^N}[\cL_{\vp}(\gA(S))]$ the expected test error on $\Dtest=\cD_\vp$ when training with $\Dtrain=\cD_\vq$ (we frequently drop the subscript $N$ if its clear from context).  The ``non-shifted'' expected test loss is then denoted $\Lsame_N(\vp)=L_N(\vp,\vp)$.  In contrast, we denote $L^*_N(\vp)=\min_{\vq \in \Delta_K} L_N(\vp,\vq)$ the test error with the best mixing ratios, and $\vq^*$ the minimizing ratios.  When $L^* < \Lsame$ and so $\vq^* \neq \vp$, this means we can benefit from mismatched training.  \textbf{Our main analysis objective is to charactarize $\vq^*$, $L^*$ and the improvement over $\Lsame$.}

We measure the mismatch benefit through the improvement in test error for a fixed data size $L^\textrm{ratio}_N=L^*_N/\Lsame_N$. Or, we measure the sample complexity $N_\epsilon(\vp,\vq) = \min\{ N : L_N(\vp,\vq)\leq\epsilon\}$ and its improvement $N_{\epsilon}^\textrm{ratio}:= N_\epsilon^*(\vp) /N_\epsilon^\textrm{same}(\vp)$.
We use the standard $O(\,\cdot\,), \Omega(\,\cdot\,), \Theta(\,\cdot\,), o(\,\cdot\,)$ notations for functions of the data size $N$ when characterizing these quantities, and hide dependence on other parameters.

\paragraph{Specifying the Learning Model} 
The expected test loss $L_N(\vp,\vq)$,  and so $\vq^*$ and the benefit of mismatch, depend on the data distributions and learning behaviour of the algorithm.  We capture these by modeling the \emph{subpoluation error function} (or per-component learning curves) $e_k(n_k)$, i.e. the error on each component $\gD_k$ when training with $n_k$ examples.  That is, for a vector of sample sizes $\vn = (n_1,\ldots,n_K) \in \mathbb{Z}_{\ge 0}^K$, denote $\boldsymbol{\cD}^\vn = (\cD_1)^{n_1}\times \dots \times (\cD_K)^{n_K}$ the distributions over samples with $n_i$ examples from each component $\cD_i$.  Then $e_k(n_k) = \mathbb E_{S\sim\boldsymbol{\cD}^\vn}[\cL_{\cD_k}\left(\gA(S) \right)]$. The scalar function $e_k(n_k)$ captures the {\em learning curve} for each component. We focus on the case when there is no interference (positive or negative) or transfer between tasks\footnote{When we say there is no interference, it is easiest to think of $\cD_i$ as having disjoint support, but we do not formally require this as we treat $\cD_i$ abstractly and only model the error functions.}, i.e.~when training one task neither helps nor hurts the others, so each $e_k$ is only a function of $n_k$. In the next two sections, we consider different learning settings, specified by different types of error functions, and characterize $\vq^*$ and $L^*$ in terms of the error functions.  In \Cref{sec:transfer} we also consider a setting with transfer between components.  

\paragraph{Datasets and Training Sequences}  In our analysis, we refer to the training budget $N$ and our learning model specifying learning based on $n_k$ examples per component $k$.  We can think of $N$ and $\vn$ as specifying the number of training examples, in which case the training complexity is a sample complexity.  Or, we can think of $N$ as indicating the number of training steps, and $n_k$ as indicating the number of steps in which an example from component $k$ is used.  In this case, training complexity is a measure of training time.  Either interpretation is valid.  But we should emphasize that we only study a dependence on {\em how many} examples are used from each component, {\em not} on the {\em order} (as in curriculum learning).

\paragraph{Learnabilities and Mixing Ratios.} We model learning as a function of the {\em number} of examples from each component, but for our analysis, it will useful to introduce the function $\bar{e}_{N,k}(\vq) =\mathbb E_{S\sim (\cD_{\vq})^n}[\cL_k(\gA(S))]$, which captures the expected error on component $k$ with mixing proportions $\vq$. We will refer to $\bar{e}_k(\vq)$ as the subpopulation error function in terms of the mixture $\vq$. Since the per-component counts $\vn$ are multinomial, we have $\bar{e}_N(\vq) = \mathbb E_{\vn \sim \text{Mult}(\vq,N)}[e(\vn)] \in \mathbb{R}^K$ and $L_N(\vp,\vq) = \langle \vp , \bar{e}_N(\vq)\rangle$. Frequently for large sample size $N$, $e(\vn), \vn \sim \text{Mult}(\vq,N)$, will concentrate around $e(\vq N)$, and we will sometimes exploit this in the analysis, or analyze for $\bar{e}_N(\vq) \approx e(\vq N)$.


\paragraph{Knowledge of Test Distribution Mixture Proportions at Test Time.} Our main motivation, and the main way to interpret our work, is addressing how could we have positive distribution shift in the mixture setting, i.e.,~how could $\Dtrain\neq \Dtest$ be better than $\Dtrain=\Dtest$, and what qualitative  changes in $\Dtrain$ make it better.  This provides guidance in understanding positive distribution shift and seeking good training distributions.   Nevertheless, there are also several realistic examples where it is conceivable the test mixing proportions are known and the training mixing proportions can be controlled or specified.  For example, if we are pretraining a large language model for a set of tasks, we might well know their frequency at test time. Even in settings where the exact mixing proportions are unknown, it suffices to estimate the mixing proportions roughly by a quick analysis of test samples. If we suspect a mixture structure in the data, we can try to build crude classifiers for the components. This can be easy for a variety of tasks. For example, in a language-related task if each mixture component is a different language, or in a memorization task, if each mixture component is a topic or area (e.g., sports, science, etc.), we can build crude topic classifiers based on a small amount of data for each topic, or perhaps unsupervised clustering of a sample, then classify a sample. We can estimate the unknown mixing proportions using this classifier. Note that the classifier does not have to be very accurate since we do not care about individual errors, just about the aggregate proportions, and a bit of an error on the proportions is fine. Similar approach is taken in \cite{ye2025mixing}, where the authors propose mixture dependent scaling laws for finding good training mixture proportion and assume that the validation data comes from an unknown mixture, which they estimate as part of their procedure in finding the good mixture proportions. 





\section{Power Law}\label{sec:orthogonalpowerlaw}

Many machine learning tasks can be captured with power law error functions. Some classic examples include linear regression or learning VC classes, both of which have error rate $\propto\frac{1}{N}$, where $N$ is the number of data samples~\citep{shalev2014understanding}. More recently, there have been many papers studying the loss curves of large language models for different tasks as a function of the compute budget or the number of training tokens through various scaling laws, which model error as having power law dependence on the number of data samples, such as the Chinchilla Scaling Law~\citep{hoffmann2022computeoptimal}, and many others \citep{kaplan2020scalinglawsneurallanguage,cherti2023scaling,ye2025mixing}.

To model these situations, we will first consider a setup where all of the $K$ tasks have subpopulation error functions that follow a simple power law in terms of the number of samples.



\begin{model}[Power Law Error Tasks]\label{model:generalpowerlaw}
    There are $K$ tasks. Each task takes data from one of the $K$ subpopulations $\cD_i$ that appear in the test distribution with probability $p_i$ and has subpopulation error functions $e_k(n_k)$ that follow a power law, i.e. $e_k(n_k) = \frac{A_k}{n_k^{\alpha_k}+B_k}$ for some $A_k>0,B_k > 0$, and $0<\alpha_k\le 1$.
\end{model}

In \Cref{thrm:main:generalpowerlaw}, we characterize the test error improvement from the positive distribution shift from optimal data mixing ratios in \Cref{model:generalpowerlaw} when the size of the training data $N$ is large.

\begin{theorem}[Optimal Data Mixing Ratios For Power Law]\label{thrm:main:generalpowerlaw}
    
    In \Cref{model:generalpowerlaw}, if for the exponents it holds that $\alpha_1 = \alpha_2 = \dots = \alpha_S < \alpha_{S+1} \le \alpha_{S+2} \le \dots \le \alpha_{K}$ for some $S$, then there exist $\varepsilon_1,\varepsilon_2>0$ such that for any test data mixing ratio $\vp$ and any $N>N_0(\{A_i,B_i,\alpha_i,p_i\}_{i=1}^{K})$ we have that the following holds 
    \begin{align}
    &\begin{aligned}\label{eq:powerlaw-q}
    q_i^* &=\frac{1}{N^{\frac{\alpha_i-\alpha_1}{\alpha_i+1}}}\left(\frac{\alpha_i p_i A_i}{\left( \sum_{i=1}^{S}(\alpha_i p_iA_i)^{\frac{1}{\alpha_1+1}}\right)^{\alpha_1+1}}\right)^{\frac{1}{\alpha_i+1}}+o\left(\frac{1}{N^{\frac{\alpha_i-\alpha_1}{\alpha_i+1}}}\right) 
    \end{aligned} \\
    &\Lsame(\vp) = \frac{1}{N^{\alpha_1}}\sum_{i=1}^{S} p_i^{1-\alpha_1} A_i + o\left(\frac{1}{N^{\alpha_1+\varepsilon_1}}\right). \label{eq:powerlaw-loss-same} \\
    &L^*(\vp) = \frac{1}{N^{\alpha_1}} \left(\sum_{i=1}^{S} (\alpha_i p_i A_i)^{\frac{1}{\alpha_i+1}} \right)^{\alpha_1} \left(\sum_{i=1}^{S} \frac{(p_iA_i)^{\frac{1}{\alpha_i+1}}}{\alpha_i^{\frac{\alpha_i}{\alpha_i+1}}} \right) \notag + o\left(  \frac{1}{N^{\alpha_1+\varepsilon_2}}\right). \label{eq:powerlaw-loss-opt}
    \end{align}
\end{theorem}

\Cref{thrm:main:generalpowerlaw} shows that in the Power Law \Cref{model:generalpowerlaw}, positive distribution shift from optimal data mixing ratios improves the prefactor of the test error dependence on the number of data samples $N$ but does not change the decay rate in terms of $N$. For the proof of \Cref{thrm:main:generalpowerlaw} and a more precise statement, including the closed form of $N_0(\{A_i,B_i,\alpha_i,p_i\}_{i=1}^{K})$, see \Cref{app:powerlaw}.

Furher, we will show that the improvement for positive distribution shift can have significant implications for making training more data efficient. To do so, we show the improvement from this positive distribution shift on the sample complexity in the case where we have one majority population and $K-1$ minority populations that all have the same power exponent $\alpha$. This will also include the test-taking example from \Cref{sec:introduction}.



\begin{corollary}[Sample Complexity Improvement From Optimal Data Mixing For General Power Law]\label{cor:main:generalpower}
    Consider \Cref{model:generalpowerlaw} with $S=K$, i.e. $\alpha_1= \dots = \alpha_K = \alpha$, $A_1=\dots = A_K = A$, and $B_1=\dots=B_k=B$ with $\vp = (p, \frac{1-p}{K-1}, \dots, \frac{1-p}{K-1})$. We have that for any $\epsilon>0$ 
    \begin{align*}
        N_{\epsilon}^{\textrm{ratio}}(\vp) & 
        \le (1-p)+2^{\frac{\alpha+1}{\alpha}}\left(\frac{p}{1-p}\right)^{\frac{1}{\alpha+1}}K^{-\frac{\alpha}{\alpha+1}}.
    \end{align*}
    Furthermore, the optimal mixing ratios are given by $q^*_{1} \propto p^{\frac{1}{\alpha+1}}$ and  $q^*_{i} \propto \left(\frac{1-p}{K-1}\right)^{\frac{1}{\alpha+1}}$ for $i\ge 2$. 
\end{corollary}

\begin{figure*}[h]
    \includegraphics[width=0.35\textwidth]{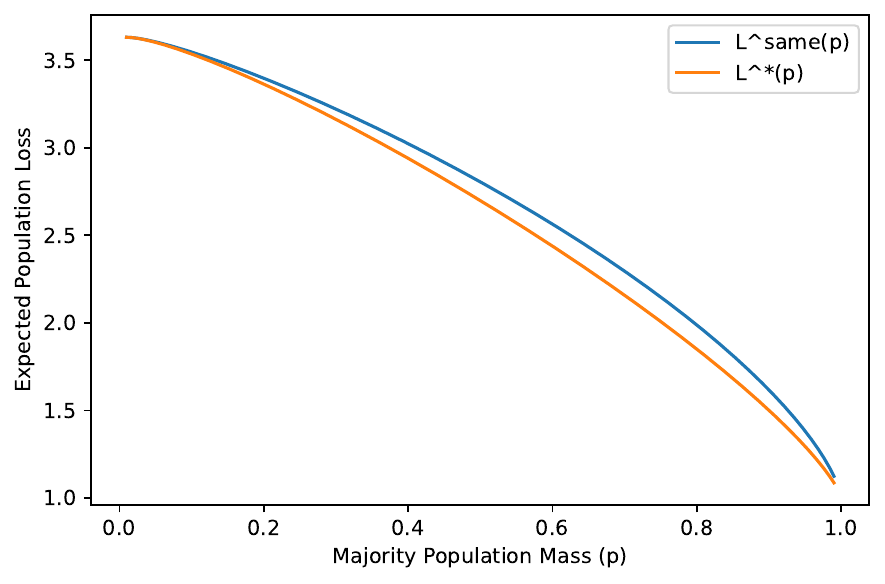}
    \includegraphics[width=0.35\textwidth]{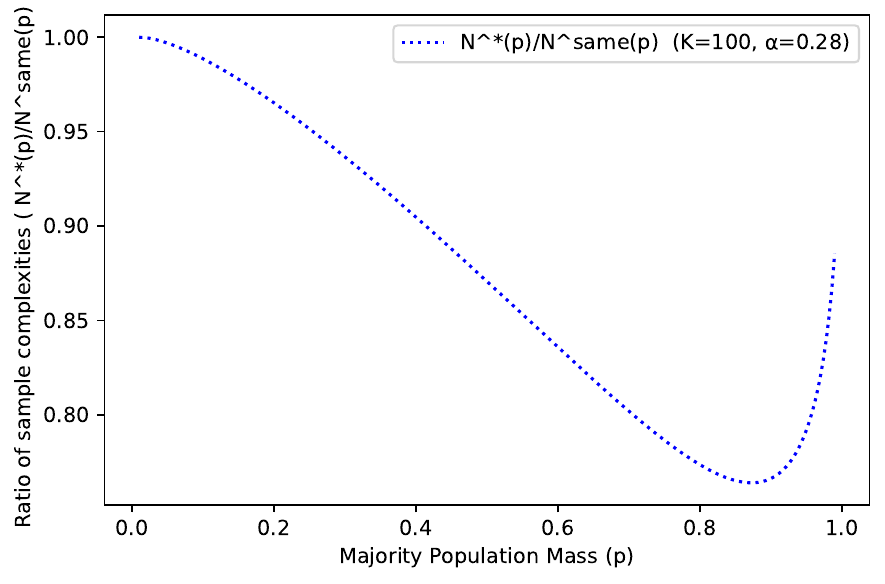}
    \centering
    \vspace{-0.15in}
    \caption{\small We consider the setup of \Cref{cor:main:generalpower} with $A=1,~\alpha=0.28$, $K=100$, and some fixed $N$.  On the left plot, we show the ``non-shifted'' expected population loss $\Lsame(\vp)$ and the optimally mixed expected population loss $L^*(\vp)$ as a function of majority population mass $p$. On the right plot, we show the ratio of sample complexities for any fixed $\epsilon>0$, $N^{\textrm{ratio}}_{\epsilon}(\vp)$ as a function of the mass of the majority population, $p$. We can see significant improvement in the sample complexity from the positive distribution shift from using optimal mixing ratio, even up to $\approx 25\%$.}\label{fig:majority-population} 
\end{figure*}




\Cref{cor:main:generalpower} demonstrates that if we have one majority population and a number of minority populations, the positive distribution shift from optimal data mixing ratio significantly improves sample complexity. For fixed $p$, if $K$ is large enough, $N^{\textrm{ratio}}(\vp)$ will be close to $N^{\textrm{ratio}}(\vp)\approx 1-p<1$, i.e. we get sample complexity improvement of up to $p$. For example, for $p=0.7$, $\alpha=0.28$, and $K=100$, for any $\epsilon>0$,  $N_{\epsilon}^{\textrm{ratio}}(\vp)\approx 0.75$, i.e. we achieve the same error with $\approx 25\%$ less data samples. We illustrate this in \Cref{fig:majority-population}. For the proof of \Cref{cor:main:generalpower}, see \Cref{app:powerlaw}. 

Furthermore, the test taking example considered in the introduction \Cref{sec:introduction} follows from \Cref{cor:main:generalpower}, by taking  $K=2$,  $\alpha=1$, and $\vp = (0.9,0.1)$ (with any $A$ and taking $B$ much smaller than $1$). In particular, this shows that the optimal studying budget allocation is $\vq^* =(0.75,0.25)$ and the improvement is $N^{\textrm{ratio}}(\vp) = 0.8$. This means that if you study for the exam with the right mixing ratio $\vq^*$, you would need to study $20\%$ less time to achieve the same score as compared to using the test mixing ratio $\vp$. Further, taking $\alpha=\frac{1}{2}$ we get the second example on \Cref{fig:majority-population}. This shows that we indeed get $\vq^* = (0.812\dots,0.188\dots)$ and $N^{\textrm{ratio}}(\vp) = 
        0.944$.

\section{Memorization Tasks}\label{sec:memorization}


Many machine learning tasks involve memorizing a number of unique atoms, such as training LLMs to answer factual questions or explaining the meaning of words, performing tabular RL, and learning transition functions in automata.
Here, the atoms correspond to different facts, answers to questions, word meanings, or states. The loss depends on what fraction of atoms the model memorized. The data distribution in this case corresponds to the training data, i.e., in the case of fact retrieval, question answer, or learning word meanings with an LLMs, the data distribution corresponds to the text used for (pre)training.

To model this, we consider a task of memorizing a number of unique atoms from a dataset of fixed size, where the test distribution is a mixture of the tasks we are trying to memorize. More explicitly, let $S$ be the set of possible atoms to memorize and let $s_1,\dots,s_k\in S$ be the $k$ atoms we are interested in memorizing. Assume that the learning rule memorizes all atoms it has seen so far, and let $M$ be the set of atoms the model has seen. Let the $i$th component (or task) be memorizing atom $s_i$. We incur error $0$ if $s_i\in M$ and error $1$ if $s_i\in M$. 

\vspace{-0.02in}
\begin{model}[Memorization Tasks]\label{model:othogonalmemorization}
    Suppose there are $K$ tasks, each of which is a memorization of a unique atom. The test distribution is a mixture of these $K$ tasks, where the $k$-th task appears with probability $p_k$. In this case the subpopulation error functions in terms of $\vn$ are given by $e_k(n_k)=\textbf{1}_{\{n_k=0\}}$.
\end{model}

\vspace{-0.02in}
The following theorem characterizes the test error improvement from the positive distribution shift from optimal data mixing ratios in the Memorization Task \Cref{model:othogonalmemorization}.

\vspace{-0.02in}
\begin{theorem}[Optimal Data Mixing Test Error Improvement For Memorization Task]\label{thrm:main:orthogonalmemorization}
    In \Cref{model:othogonalmemorization}, for all $\vp \in \Delta^{K-1}$ with $p_1 \ge p_2 \ge \cdots \ge p_K$, the expected loss when training on $n$ samples is given by
    \vspace{-0.1in}
    \begin{align}
        &\Lsame(\vp) = \sum_{k=1}^{K} p_k (1 - p_k)^N \\
        & L^*(\vp) = (K_N(\vp) - 1) \delta_N(\vp) + \sum_{k = K_N(\vp) + 1}^{K} p_k,
    \end{align}
    where $\delta_N(\vp) \in [p_{K_N(\vp)+1},\; p_{K_N(\vp)})$
    and
    $K_N(\vp)$ is defined as follows:
    \begin{equation*}
        K_N(\vp) := \max\left\{s \le K : \sum_{k=1}^{s-1} (1 - (p_s/p_k)^{\frac{1}{K-1}}) < 1 \right\}.
    \end{equation*}
\end{theorem}

\vspace{-0.05in}
To understand the magnitute of the test error improvement in \Cref{thrm:main:orthogonalmemorization}, we will assume that the test proportions $\vp$ follow a power law $p_k = \Theta(k^{-\alpha})$ for some $\alpha>1$ and that the number of tasks to memorize $K$ is larger than the size of the training set $N$. In this case, we show that the improvement from positive distribution shift \Cref{thrm:main:orthogonalmemorization} improves even the test error scaling in terms of $N$. For the proof of \Cref{thrm:main:orthogonalmemorization}, see \Cref{app:orthogonalmemorization}.

\begin{corollary}[Test Error Improvement For Memorization Taks with Power Law Test Mixing Ratios]\label{cor:main:orthogonalmemorization}
    If $p_k = \Theta(k^{-\alpha})$ for some $\alpha > 1$
    and $K = \Omega(N)$, then
    \vspace{-0.05in}
    \begin{align*}
        \Lsame(\vp) = \Theta(N^{-1+\frac{1}{\alpha}}), \qquad
        L^*(\vp) = \Theta(N^{-\alpha+1}).
    \end{align*}

\end{corollary}

\vspace{-0.1in}
For example, when $\alpha = 1.5$, we have $\Lsame(\vp) = \Theta(N^{-1/3})$ and $L^*(\vp) = \Theta(N^{-1/2})$. For the proof of \Cref{cor:main:orthogonalmemorization}, see \Cref{app:orthogonalmemorization}.

\section{Connection to Skill Composition}\label{sec:composition}

Training language models, especially for reasoning tasks including mathematical reasoning, naturally requires the models to learn multiple independent simple skills, and then compose these skills when solving a problem. In this setting, the natural distribution of problems induces a natural distribution of skills at test time. Should we always train the model on the same distribution of skills as the test skill distribution?
This is not a mixture-model per-se: each instance is a problem and thus includes many skills, and so the test distribution is {\em not} a mixture distribution over problems requiring different skills.  Nevertheless, in this section, we show that the answer to this question closely follows for the mixture proportions analysis in the previous sections.

\paragraph{A Stylized Model for Compositional  Reasoning.} To model the above skill composition scenario, we consider the following model:  we assume a problem distribution $\gD_{\mathrm{P}}$, where each problem in the distribution requires sequentially applying $k$  skills $(g_1, g_2, \dots, g_k)$ to an input $x$. 
That is, $a^*_0 = x$, $a^*_i = g_i(a^*_{i-1})$ for all $i = 1, \dots, k$.
We denote a problem by $(x, g_1, g_2, \dots, g_k)$, and let the set of all skills be~$\gG$.

A language model $h$ is trained to solve these problems with Chain-of-Thought (CoT) reasoning.
After reading the problem $(x, g_1, g_2, \dots, g_k)$, it attempts to apply the skills in order, generating
\vspace{-0.1in}
\begin{align*}
&a_0 = x,  \\
&a_i \sim P_h(a_i \mid x, g_1, \dots, g_k, a_0, \dots, a_{i-1}),~~i=1,\dots,k,
\end{align*}
where $P_h$ is the sequence distribution induced by the language model $h$.  
The test accuracy is defined as
$\mathrm{acc}_{\mathrm{test}}(h) = P_{(x, g_1, \dots, g_k) \sim \gD_{\mathrm{P}}}[a = a^*]$,
where $a \sim P_h(a \mid x, g_1, \dots, g_k)$ and $a^*$ is the ground-truth output.  
Expanding this gives
\begin{align*}
    \mathrm{acc}_{\mathrm{test}}(h)
    &= \E_{\gD_{\mathrm{P}}}[P_h(a^* \mid x, g_1, \dots, g_k)] \\
    &= \E_{\gD_{\mathrm{P}}}
    \left[\prod_{i=1}^{k} P_h(a^*_i \mid x, g_1, \dots, g_k, a^*_0, \dots, a^*_{i-1}) \right].
\end{align*}

\vspace{-0.1in}
To connect with the settings of learning multiple tasks discussed in previous settings, we now introduce two approximations, each of which amounts to an independence assumption:

\vspace{-0.1in}
\begin{enumerate}
    \item \textbf{Step Locality.} The probability of applying the $i$-th skill correctly depends only on the skill identity, not on the specific input or the surrounding context:  
   $P_h(a^*_i \mid x, g_1,\dots,g_k, a^*_0,\dots,a^*_{i-1}) \approx \tilde P_h(g_i)$,
   where $\tilde{P}_h(g_i)$ is a function that only depends on~$g_i$.
   This treats skill applications as conditionally independent across steps, once the skill type is fixed.
   \item \textbf{Skill Independence.} The skills $(g_1,\dots,g_k)$ in $\gD_{\mathrm{P}}$ are approximately independent in the problem distribution $\gD_{\mathrm{P}}$:
   $P_{\gD_{\mathrm{P}}}(x,g_1,\dots,g_k) \approx P_{\gD_{\mathrm{P}}}(x) \cdot \prod_{i=1}^k P_{\gD_{\mathrm{P}}}(g_i)$.
\end{enumerate}
\vspace{-0.1in}
Under these assumptions, the accuracy simplifies to
\begin{align*}
    \mathrm{acc}_{\mathrm{test}}(h)
    \approx \E_{\gD_{\mathrm{P}}}\left[\prod_{i=1}^{k} \tilde{P}_h(g_i) \right]
    &\approx \prod_{i=1}^{k} \E_{g \sim P_{\gD_{\mathrm{P}}}}[\tilde{P}_{h}(g)] \\
    &= \bar{p}(h)^k,
\end{align*}
where $\bar{p}(h) := \sum_{g \in \gG} P_{\gD_{\mathrm{P}}}(g) \tilde P_h(g)$.
That is, each skill application succeeds independently with probability $\bar p(h)$, which we may view as the model's \emph{per-skill accuracy} averaged under the test distribution.  
Solving a length-$k$ reasoning chain requires $k$ independent successes, so the overall accuracy scales as $\bar p(h)^k$.  

Now we take a closer look at the per-skill accuracy~$\bar{p}(h)$.
This connects directly to our previous multi-task learning framework: each skill $g \in \gG$ corresponds to a distinct task, and the mixing proportion $P_{\gD_{\mathrm{P}}}(g)$ represents the natural frequency with which skill $g$ appears in mathematical problems. 
In this sense, we can set $\vp \in \Delta_{|\gG|-1}$ as a vectorized version of $P_{\gD_{\mathrm{P}}}$, and define $\cL_{\vp}(h) = 1 - \bar{p}(h)$ as the per-skill test error.
Under our approximations, maximizing the overall test accuracy $\mathrm{acc}_{\mathrm{test}}(h) \approx \bar{p}(h)^k$ is equivalent to minimizing the per-skill test error $\cL_{\vp}(h)$ under the natural frequency of each skill.

Is it always good to train a model on the same per-skill distribution? From the insights we obtained from the previous sections, we see that  the best strategy may be to shift the per-skill distribution in a proper way.  While the previous sections studied this for abstract learning rules, we will demonstrate this empirically in the next part.


\paragraph{Transformer Experiments.}
We consider a concrete synthetic task on skill composition. There are $M$ skills, where the $i$-th skill is a function $g_i$ that maps a number from $\Omega := \{0, \dots, 9\}$ to $\Omega$. Each skill has a unique English ID.
Assume that all these skills are randomly sampled: the IDs are uniformly random from a ID set, and each $g_i$ is uniformly random among all possible functions that map from $\Omega$ to $\Omega$.
At inference time, a set of $k$ skills $g_{i_1}, \dots, g_{i_k}$ are sampled IID following a power law with exponent $\alpha = 1.5$. That is, $\Pr[i_k = i] \sim i^{-1.5}$ for $i = 1, \dots, M$.
The language model is prompted with the IDs of these skills and a number $x \in \Omega$: ``\verb|[x] -> [skill ID 1] -> [skill ID 2] ->| $\cdots$\verb| -> [skill ID k]|''. The model is expected to output the result after function composition: $y = g_{i_k}(g_{i_{k-1}}(\cdots g_{i_1}(x) \cdots))$.

\begin{figure*}[t]
    \vspace{-0.15in}
    \includegraphics[width=0.32\textwidth]{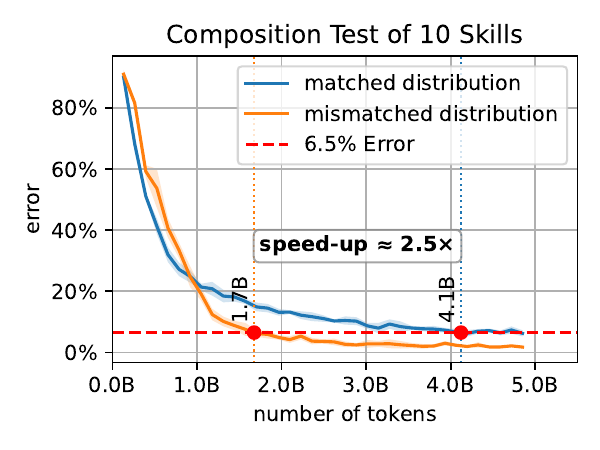}
    \includegraphics[width=0.32\textwidth]{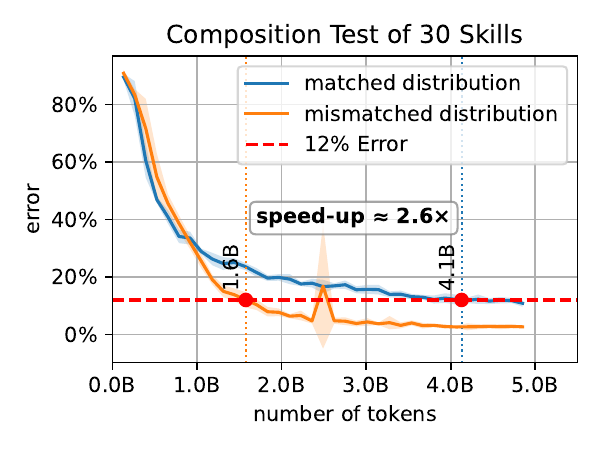}
    \includegraphics[width=0.32\textwidth]{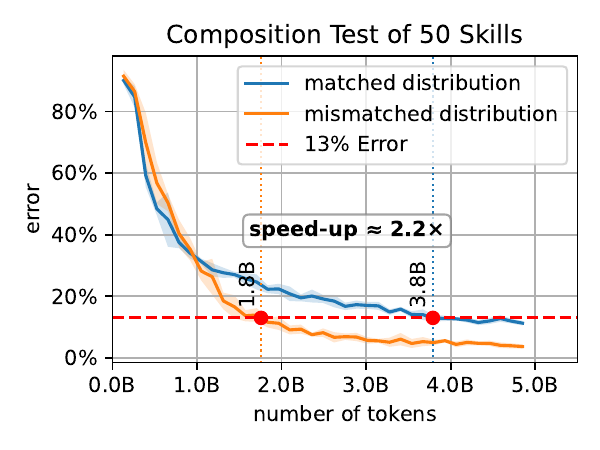}
    \centering
\vspace{-0.15in}
    \caption{
        \small
        Mismatched distribution improves the test accuracy of a language model in solving a synthetic skill composition task (\Cref{sec:composition}). During test, the model is asked to compose several functions, sampled following a power law. Instead of training directly on this task (blue curve), mixing with another task that uniformly samples the functions improves the final accuracy (orange curve).
        Curves are averaged over $5$ random seeds.
    } \label{fig:skill-comp}
\vspace{-0.1in}
\end{figure*}

Let $\Dtest$ be the distribution of the above prompt and a CoT calculating the correct answer, with $M = 10^5$, $k$ sampled uniformly from $10$ to $50$.
Is the best strategy just training on the same distribution ($\Dtrain = \Dtest$)?
Inspired by our calculation for the memorization task above, properly adjusting the occurrence probability for each skill may lead to better test accruacy.
To demonstrate this, we construct another distribution $\cD_{\mathrm{uniform}}$ consisting of strings in the form of ``\verb|[x] [skill ID] = [expected output]|'', where the skill ID and input number are uniformly sampled.
In~\Cref{fig:skill-comp}, we conduct experiments with a model with GPT-2 architecture and $\sim$50M parameters. We show that training with $\Dtrain = 30\% \cdot \cD_{\mathrm{uniform}} + 70\% \cdot \Dtest$ significantly outperform training with $\Dtest$, with around $2.5\times$ speedup in sample efficiency.
We defer the experiment details to \Cref{app:experiment}.



\section{Transfer Learning}\label{sec:transfer}




In this section, we also consider a setting where the tasks are not independent, and there is transfer. Generally, with transfer and in multitask learning, the error on the $k$-th task is affected by the number of samples on other tasks, that is, the error on task $k$ decreases if we hold $n_k$ constant but increase the number of samples on related tasks. In the most typical transfer learning setups studied in the literature, such as multi-task learning of linear classifiers over linear representation with feature learning \citep{baxter2011inductive, Maurer2009TransferBF, pontil2013excess, aliakbarpour2024metalearning} and multi-task learning with shared sparsity  \citep{wang2016distributed, wang2017distributed}, the transfer effect is captured by the following model, with a slight extension of our framework.

\begin{model}[Standard Transfer Learning Model]\label{model:standardtransfer}
    There are $K$ subpopulations, each of which appears in the test distribution with proportion $p_k$. We extend our framework to allow the subpopulation error functions to depend on all of $\vn$. Then, for the standard transfer learning model, let $e_k(\vn) = \frac{A_{0,k}}{(n_1+\dots+n_k)^{\alpha_k}+B_{0,k}}+\frac{A_{1,k}}{n_k^{\alpha_k}+B_{1,k}}$, for some $A_{0,k},A_{1,k},B_{0,k},B_{1,k}>0$ and $0<\alpha_k \le 1$. 
\end{model}

 For example, in multi-task learning of shared sparsity \citep{wang2017distributed}, the error bound takes this form with $\alpha_1=\dots=\alpha_K=1$. Interestingly, it turns out that the behavior of the test error in these typical transfer learning settings is the same with respect to the mixture proportions as for the independent component setting, and so our analysis is actually applicable to this non-independent setting as well. The Standard Transfer Learning \Cref{model:standardtransfer} is equivalent to the setup of Power Law Tasks \Cref{model:generalpowerlaw} in the sense that we can understand the optimal data mixing ratio $\vq^*$ and the error improvement of the Standard Transfer Learning model from a specific instance of the Power Law \Cref{model:generalpowerlaw}. Namely, the transfer term in each of the subpopulation loss functions can be decomposed into a transfer error term and a specific task error term $e_k(\vn)=e_k^{\textrm{transfer}}(\vn)+e_k^{\textrm{spec}}(\vn)$, where $e_k^{\textrm{transfer}}(\vn)=\frac{A_{0,k}}{(n_1+\dots+n_k)^{\alpha_k}+B_{0,k}}$ is independent of the distribution of samples across different tasks, and $e_k^{\textrm{spec}}(\vn)=e_k^{\textrm{spec}}(n_k)=\frac{A_{1,k}}{n_k^{\alpha_k}+B_{1,k}}$ only depends on $n_k$. Therefore, the transfer error term $e_k^{\textrm{transfer}}(\vn)$ in each of the subpoluation error functions will only offset the final expected loss $L(\vp ,\vq)$ by $\sum_{i=1}^{K}p_i \frac{A_{0,k}}{N^{\alpha_k}+B_{0,k}}$, which only depends on the total number of samples $N$. On the other hand, the specific task error terms $e_k^{\textrm{spec}}(n_k)$ can be thought of as independent (i.e. without transfer) tasks and will behave the same as in \Cref{model:generalpowerlaw}. So, for the Standard Transfer Learning \Cref{model:standardtransfer}, the optimal data mixing ratio $\vq^*$ and the expected test losses $L^*(\vp)$ and $\Lsame(\vp)$ are given by \Cref{eq:powerlaw-q}, \Cref{eq:powerlaw-loss-same} and \Cref{eq:powerlaw-loss-opt} respectively in \Cref{thrm:main:generalpowerlaw} with $A_k$ being replaced by $A_{1,k}$.

More complex transfer structures, as is likely usually the case in practice, could lead to even stronger positive distribution shifts, depending on how one task informs the other tasks.
In this paper we focus on showing how positive distribution shift can arise {\em even without} such transfer and understanding purely the effects of mixing proportions.  Understanding positive distribution shift more broadly in transfer settings requires specific considerations about the specific form and source of transfer, and we indeed hope to work on transfer learning being described in these terms.  We emphasize that we are not aware of this type of analysis for transer learning.  This is different from typical descriptions of transfer learning, where data from an alternate or surrogate task is seen as a compromise replacement for additional data from the target task.

\section{It's Almost Always Better to Mismatch}\label{sec:general-case}

So far, we have shown the existence of and quantified the positive distribution shift coming from mistmatched test and train data mixing ratios for the cases of power law tasks in \Cref{sec:orthogonalpowerlaw}, memorization tasks in \Cref{sec:memorization}, and standard transfer learning  in \Cref{sec:transfer}. In this section, we will show that a positive distribution shift coming from the mismatched data mixing ratio almost always exists, i.e., it is almost always better to mismatch the training and test distributions: $\vq^*\neq \vp$ and $L^*(\vp,\vq^*)<\Lsame(\vp)$.  

More precisely, we will show that if there is no positive distribution shift, either the test data mixing ratio is on a measure zero set of the simplex or the subpopulation error functions $e_k(n_k)$ have to be all constant functions, which is meaningless. We show this in \Cref{cor:main:generalcase-orthogonal}.

Let $\Delta_+^{K-1} := \left\{ \vp \in \R^K : \vp > 0,\; \abs{\vp} = 1 \right\}$ be the probability simplex and its interior, respectively, where $\abs{\vp} := \sum_{k=1}^{K} p_k$.
We define $f_k(\vp)$ by extending the domain of each $\bar{e}_k(\vp)$ to the set of non-zero, non-negative vectors $\R^K_{\ge 0} \setminus \{\vzero\}$ by defining $f_k(\vp) := \bar{e}_k( \frac{\vp}{\abs{\vp}})$.
We further define $\Lsame(\vp) := \sum_{k=1}^{K} p_k f_k(\vp)$, which extends the definition of $\Lsame$ to the set of non-zero, non-negative vectors $\R^K_{\ge 0} \setminus \{\vzero\}$.
\begin{condition}[Conservation Condition]\label{cond:gradient}
    For all $\vp \in \R^K_{\ge 0} \setminus \{\vzero\}$,
    $(f_1(\vp), \dots, f_K(\vp)) = \nabla \Lsame(\vp)$.
\end{condition}

\begin{theorem}[Positive Distribution Shift Almost Always Exists For Data Mixing]\label{thm:main}
    For any set of subpopulations $\gD_1, \dots, \gD_K$ and any learning algorithm $\gA$, either \Cref{cond:gradient} holds, or there exists a zero-measure set $U$ on $\Delta_{K-1}$ such that for all $\vp \in \Delta_{K-1} \setminus U$,
    $\Lshift^*(\vp) < \Lsame(\vp)$.
\end{theorem}

\Cref{thm:main} shows that either $\vp$ is  on a measure zero set $U$ on $\Delta_{K-1}$ or the Conservation \Cref{cond:gradient} must hold. Next, we show that if the tasks are independent, then the Conservation \Cref{cond:gradient} 
holds only if all of the subpopulation error functions are constants.

\begin{lemma}[Independent Tasks]\label{lemma:main:orthogonaltasks}
    If $K \ge 3$,
    and if for all $k \in [K]$,
    $f_k(\vp) = g_k(\frac{p_k}{\abs{\vp}})$ for some function $g_k$,
    then \Cref{cond:gradient} holds if and only if $g_k$'s are all constant functions. 
\end{lemma}

\Cref{thm:main} and \Cref{lemma:main:orthogonaltasks} together show that positive distirbution shift always exists,
unless all the subpopulation error functions are constant. 

\begin{corollary}[Positive Distribution Shift Always Exists]\label{cor:main:generalcase-orthogonal} For any set of $K\ge 3$ subpopulations $\cD_1,\dots,\cD_K$ and any learning algorithm $\gA$, 
if there exists subpopulation $k\in[K]$ such that its error function $e_k$ is not a constant functions over $[N]$ where $N$ is the number of total samples
then there exists a measure zero set $U$ on $\Delta_{K-1}$ such that for all $\vp \in \Delta_{K-1} \setminus U$ positive distribution shift from data mixing exists in the sense that there is $\vq^*\neq p$ for which $\Lshift(\vp,\vq)=L^*(\vp) < \Lsame(\vp)$.
    
\end{corollary}

For the proofs of \Cref{thm:main}, \Cref{lemma:main:orthogonaltasks}, and \Cref{cor:main:generalcase-orthogonal}, see \Cref{app:generalcase}.

\section{Related Works}

\paragraph{Distribution Shift That is Not Harmful.} The benefits of mismatching the training and test distribution has already been in studied in some settings. \citet{gonzalez2015mismatched} demonstrate positive distribution shift in an entirely different setting, namely linear regression problems with generic mismatched training and test distributions. Unlike in our paper, they do not restrict to changing the train distribution only through data mixing, and generally only show the existence of positive distribution shift in linear rergression problems.  They are able to charactarize the optimal shift explicitly only in very special cases. \citet{canatar2021out} show how to numerically optimize the training distribution in high-dimensional kernel regression problems. However, they do not characterize the positive distribution shift, but rather only show how to numerically find it for kernel regression. They do not restrict the test distribution to one coming from a data mixture. Since our focus is on mixing proportions, neither of these investigations fit our framework. 

\paragraph{Data Mixture Selection.} There are a number of empirical papers on finding the optimal data mixture ratios.  \cite{xie2023doremioptimizingdatamixtures} and \cite{liu2025regmix} train a smaller proxy model to find good mixing proportions and use those for training a large model. \cite{albalak2023efficientonlinedatamixing} develops a multi-arm bandit algorithm for online optimization of mixing proportions. \cite{ye2025mixing} and \cite{shukor2025scalinglawmixing} propose new scaling laws that depend on the mixture coefficients for determining the optimal mixing ratio for a target domain. \cite{jiang2025adaptive}adaptively select the mixing proportions based on the scaling laws for each domain separately.  The papers consider both optimizing mixing proportions so as to optimize test performance on a different target distribution, and in some papers \citep{xie2023doremioptimizingdatamixtures, ye2025mixing, albalak2023efficientonlinedatamixing,  jiang2025adaptive, shukor2025scalinglawmixing} also to optimize test performance on a target which is a mixture of the training components with some fixed target proportions---as in our setup.  All the papers focus on methods for empirically optimizing training proportions, rather than understanding and charactarizing the phenomena.  More importnatly, in the settings considered in these papers there is significant transfer between components, which no doubts dominates the positive distribution shift---we consider a `pure' setting with orthogonal tasks and no transfer to emphasize how Positive Distribution Shift can occur even in such a setting.

\section{Summary}



In this paper, we investigate one of the several ways in which distribuion shift can be \emph{positive}, in particular focusing on how positive distribution shift can happen due to mismatched training and test mixture proportions in the setting where the test distribution is a mixture of $K$ tasks or components. We specifically consider independent tasks to understand how mismatched proportions themselves lead to positive distribution shift. We show that in this setting, the optimal training distribution is never equal to the test distribution, except for a measure zero set of test distributions or for those satisfying a conservation property that does not generally hold. Furthermore, we consider different per-components learning curves and the possibility of transfer and in all of these cases we characterize the optimal training mixture and the improvement in sample complexity coming from positive distribution shift.

\bibliography{main}
\bibliographystyle{plainnat}

\clearpage
\appendix
\thispagestyle{empty}

\onecolumn

\section{Proofs of Error Rate Improvements}\label{app:errorimprovement}



\subsection{General Power Law Tasks}\label{app:powerlaw}

\begin{definition}[Approximate Subpopulation Error Function]\label{def:app:approxpowerlaw}
For Power Law \Cref{model:generalpowerlaw}, let $f_k(\vq)$ be \emph{approximate subpopulation error function} defined as 
\begin{align*}
    f_k(\vq)  &= \frac{A_k}{(q_kN)^{\alpha_k}+B_k}.
\end{align*}
We define the \emph{approximate expected population loss} as 
\begin{align}\label{eq:approximate-loss}
    \tilde{L}(\vp,\vq) &=\sum_{i=1}^{K} p_i f_i(\vq) = \sum_{i=1}^{K}p_i \frac{A_i}{(q_iN)^{\alpha_i}+B_i}.
\end{align}
\end{definition}

First, we show that for Power Law \Cref{model:generalpowerlaw} and large number of samples $N$, it is sufficient to optimize over the approximate expected population loss to find $\vq*$ up to error of the order $\frac{1}{N}$.

\begin{proposition}[Sufficient to Consider Expectation]\label{lemma:app:sufficientfk}
    For the approximate error function $f_k(\vq)$ in \Cref{def:app:approxpowerlaw}, we have that when $N q_k$ is large enough,
    \begin{align*}
        |f_k(\vq)-\bar{e}_k(\vq)|\le \frac{320A_k}{B_k} \cdot \frac{1}{(Nq_k)^2}+\frac{\alpha_k A_k }{(Nq_k)^{\alpha_k+\frac{1}{4}}}  + \frac{\alpha_kA_k}{(Nq_k)^{\alpha_k+\frac{1}{2}}}.
    \end{align*}
\end{proposition}

\begin{proof}[Proof of \Cref{lemma:app:sufficientfk}]
Let $g_k(x) = \frac{A_k}{x^{\alpha_k}+B_k}$. Note that for $n_k\sim \textrm{Binom}(N,q_k)$ we have that $\mu = \E[n_k]=Nq_k$. So, we have that $f_k(\vq)= g_k(\mu)$ and $\bar{e}_k(\vn) = \E[g_k(n_k)]$. Note also that on $(0,\infty)$, $g_k(x)$ is twice differentiable with 
\begin{align*}
g_k'(x) &= -\frac{A_k\,\alpha_k\,x^{\alpha_k-1}}{\bigl(x^{\alpha_k}+B_k\bigr)^2} \\
g_k''(x) &= \frac{A_k\,\alpha_k(1-\alpha_k)\,x^{\alpha_k-2}}{\bigl(x^{\alpha_k}+B_k\bigr)^2}+\frac{A_k\,\alpha_k^2\,x^{2\alpha_k-2}}{\bigl(x^{\alpha_k}+B_k\bigr)^3} = \frac{A_k \alpha_k((1-\alpha_k) B_k + (\alpha_k + 1) x^{\alpha_k} ) x^{\alpha_k-2}}{(x^{\alpha_k} + B_k)^3}
\end{align*}
First, we decompose $\E[g_k(n_k)-g_k(\mu)]$ into two parts:
\begin{align*}
    \E[g_k(n_k)-g_k(\mu)] &= \underbrace{\E\left[(g_k(n_k)-g_k(\mu)) \onec{|n_k-\mu|< \mu^{3/4}}\right]}_{=: \delta_1} + \underbrace{\E\left[(g_k(n_k)-g_k(\mu)) \onec{|n_k-\mu|\ge \mu^{3/4}}\right]}_{=: \delta_2}.
\end{align*}
For $\delta_2$, by the Multiplicative Chernoff bound we have that if $\mu>1$ 
\begin{align*}
    P\left(|n_k-\mu|\ge \mu^{\frac{3}{4}} \right) \le 2\exp(-\frac{\sqrt{\mu}}{2}).
\end{align*}
Therefore, we have that
\begin{align*}
    \E[(g_k(n_k)-g_k(\mu))\onec{|n_k-\mu|\ge \mu^{3/4}}] \le2\exp(-\frac{\sqrt{\mu}}{2}) \frac{A_k}{B_k} \le \frac{320A_k}{B_k} \frac{1}{\mu^2}.
\end{align*}
where we used the fact that $|g_k(n_k)-g_k(\mu)|\le \max_{x \ge 0} g_k(x) \le \frac{A_k}{B_k}$.

For $\delta_1$, by Taylor's Theorem, there exists $\xi\in (n_k,\mu)$ (or $(\mu,n_k)$) so that 
\begin{align*}
    g_k(n_k) = g_k(\mu) + g_k'(\mu) (n_k-\mu) + \frac{1}{2}g_k''(\xi) (n_k-\mu)^2.
\end{align*}
So we have that
\begin{align*}
    \labs{\delta_1}
    &= \labs{\E\left[g_k'(\mu) (n_k-\mu) \onec{|n_k-\mu|< \mu^{3/4}}\right]
    + \E\left[\frac{1}{2}g_k''(\xi) (n_k-\mu)^2 \onec{|n_k-\mu|< \mu^{3/4}}\right]} \\
    &\le \labs{\E\left[\abs{g_k'(\mu)} \cdot \abs{n_k-\mu} \cdot \onec{|n_k-\mu|< \mu^{3/4}} \right]}
    + \frac{1}{2}\left(
        \sup_{x \in (\mu - \mu^{3/4}, \mu + \mu^{3/4})} \labs{g_k''(x)}
    \right) \cdot \mu^{\frac{3}{2}} \\
    &\le 
    \frac{\alpha_k A_k }{\mu^{\alpha_k+1}} \mu^{\frac{3}{4}} + \frac{\alpha_k(1-\alpha_k)A_k}{\mu^{\alpha_k+2}} \mu^{\frac{3}{2}}+\frac{\alpha_k^2A_k}{\mu^{\alpha_k+2}} \mu^{\frac{3}{2}}\\
    & \le \frac{\alpha_k A_k }{\mu^{\alpha_k+\frac{1}{4}}}  + \frac{\alpha_kA_k}{\mu^{\alpha_k+\frac{1}{2}}}.
\end{align*}

Where we used the following bound on the supremum of $g''_k(x)$: $\left(
        \sup_{x \in (\mu - \mu^{3/4}, \mu + \mu^{3/4})} \labs{g_k''(x)}
    \right)\le \frac{A_k \alpha_k \left( (1-\alpha_k)B_k+(\alpha_k+1)(\mu-\mu^{3/4})^{\alpha_k}\right)(\mu-\mu^{3/4})^{\alpha_k-2}}{\left((\mu-\mu^{3/4})^{\alpha_k}+B_k\right)^{3}}\le \frac{A_k \alpha_k^2 \mu^{\alpha_k} \mu^{\alpha_k-2} }{\mu^{3\alpha_k}}$, as long as $\mu-\mu^{3/4} \ge B_k$ and $(\mu-\mu^{3/4})^{\alpha_k} \ge B_k$, which happens for $Nq_k \ge 2\max\{B_k, B_k^{\alpha_k} \}$.

Putting all these together proves the proposition.
\end{proof}

\begin{proposition}[Optimum of the Approximate Power Law]\label{lemma:app:soltoapprox}
    Let $\tilde{q}^*$ be the minimum of the approximate population loss defined in \Cref{eq:approximate-loss}. For $N>N_0(p_i,A_i,B_i,\alpha_i)$, we have that then 
    \begin{align*}
     \tilde{q}_i^* &=\frac{1}{N^{\frac{\alpha_i-\alpha_1}{\alpha_i+1}}}\left(\frac{(\alpha_i p_i A_i)}{\left( \sum_{i=1}^{S}(\alpha_i p_iA_i)^{\frac{1}{\alpha_1+1}}\right)^{\alpha_1+1}}\right)^{\frac{1}{\alpha_i+1}}+o\left(\frac{1}{N^{\frac{\alpha_i-\alpha_1}{\alpha_i+1}}}\right)  
    \end{align*}
    
    \begin{align}
        \tilde{L}(\tilde{\vq}^*) &= \frac{1}{N^{\alpha_1}} \left(\sum_{i=1}^{S} (\alpha_i p_i A_i)^{\frac{1}{\alpha_i+1}} \right)^{\alpha_1} \left(\sum_{i=1}^{S} \frac{(p_iA_i)^{\frac{1}{\alpha_i+1}}}{\alpha_i^{\frac{\alpha_i}{\alpha_i+1}}} \right)+ O\left(  \frac{1}{N^{\alpha_1+\frac{3\alpha_1^2}{2\alpha_1+2}}}\right).
    \end{align}
\end{proposition}

\begin{proof}[Proof of \Cref{lemma:app:soltoapprox}]
    We will take $N$ large enough so that we force $\tilde{q}_i^*\neq 0$, which we do as follows. 
    First take 
    \begin{align*}
        r_i = \frac{1}{N^{\frac{\alpha_i-\alpha_1}{\alpha_i+1}}}\left(\frac{(\alpha_i p_i A_i)}{\left( \sum_{i=1}^{S}(\alpha_i p_iA_i)^{\frac{1}{\alpha_1+1}}\right)^{\alpha_1+1}}\right)^{\frac{1}{\alpha_i+1}}
    \end{align*}
    Take 
    \begin{align*}
        \bar{q}_1 &= \left(\frac{(\alpha_1 p_1 A_1)}{\left( \sum_{i=1}^{S}(\alpha_i p_iA_i)^{\frac{1}{\alpha_1+1}}\right)^{\alpha_1+1}}\right)^{\frac{1}{\alpha_1+1}} - \sum_{i=S+1}^{K} \frac{1}{N^{\frac{\alpha_i-\alpha_1}{\alpha_i+1}}}\left(\frac{(\alpha_i p_i A_i)}{\left( \sum_{i=1}^{S}(\alpha_i p_iA_i)^{\frac{1}{\alpha_1+1}}\right)^{\alpha_1+1}}\right)^{\frac{1}{\alpha_i+1}}\\
        \bar{q}_i &= r_i ~\text{ for $i>1$}.
    \end{align*}
    This way $\sum_{i=1}^{K}\bar{q}_i = 1$. Take $N$ large enough so that $\bar{q}_1\in (0,1)$, i.e. 
    \begin{align}\label{eq:first-N-cond}
        N>\left(2\left(\frac{(\alpha_1 p_1 A_1)}{\left( \sum_{i=1}^{S}(\alpha_i p_iA_i)^{\frac{1}{\alpha_1+1}}\right)^{\alpha_1+1}}\right)^{\frac{-1}{\alpha_1+1}} \left(\sum_{i=S+1}^{K}\left(\frac{(\alpha_i p_i A_i)}{\left( \sum_{i=1}^{S}(\alpha_i p_iA_i)^{\frac{1}{\alpha_1+1}}\right)^{\alpha_1+1}}\right)^{\frac{1}{\alpha_i+1}} \right)\right)^{\frac{\alpha_{S+1}+1}{\alpha_{S+1}-\alpha_1}}
    \end{align}
    suffices because $\frac{1}{N^{\frac{\alpha_{S+1}-\alpha_1}{\alpha_{S+1}+1}}}\ge \frac{1}{N^{\frac{\alpha_i-\alpha_1}{\alpha_i+1}}}$ for all $i\ge S+1$.
    Note that for these $\bar{q}_i$, we have that for all $i>2$
    \begin{align*}
        f_i(\bar{\vq})\le \frac{1}{N^{\alpha_i \frac{1+\alpha_1}{1+\alpha_i}}} A_i\left(\frac{(\alpha_i p_i A_i)}{\left( \sum_{i=1}^{S}(\alpha_i p_iA_i)^{\frac{1}{\alpha_1+1}}\right)^{\alpha_1+1}}\right)^{-\frac{\alpha_i}{\alpha_i+1}}.
    \end{align*}
     For $i=1$, we have that $\bar{q}_1 \ge \frac{1}{2} \left(\frac{(\alpha_1 p_1 A_1)}{\left( \sum_{i=1}^{S}(\alpha_i p_iA_i)^{\frac{1}{\alpha_1+1}}\right)^{\alpha_1+1}}\right)^{\frac{1}{\alpha_1+1}}$, so 
    \begin{align*}
        f_1(\bar{\vq}) \le A_1 2^{\alpha_1} \left(\frac{(\alpha_1 p_1 A_1)}{\left( \sum_{i=1}^{S}(\alpha_i p_iA_i)^{\frac{1}{\alpha_1+1}}\right)^{\alpha_1+1}}\right)^{\frac{-\alpha_1}{\alpha_1+1}}.
    \end{align*}
    Therefore, we have that for the approximate expected population loss
    \begin{align*}
        \tilde{L}(\bar{\vq})&\le \frac{1}{N^{\alpha_1}} p_1A_1 2^{\alpha_1} \left(\frac{(\alpha_1 p_1 A_1)}{\left( \sum_{i=1}^{S}(\alpha_i p_iA_i)^{\frac{1}{\alpha_1+1}}\right)^{\alpha_1+1}}\right)^{\frac{-\alpha_1}{\alpha_1+1}} + \sum_{i=2}^{K}p_i \frac{1}{N^{\alpha_i \frac{1+\alpha_1}{1+\alpha_i}}} A_i\left(\frac{(\alpha_i p_i A_i)}{\left( \sum_{i=1}^{S}(\alpha_i p_iA_i)^{\frac{1}{\alpha_1+1}}\right)^{\alpha_1+1}}\right)^{-\frac{\alpha_i}{\alpha_i+1}} \\
        &\le \frac{2^{\alpha_1}}{N^{\alpha_1}} \left(\sum_{i=1}^{K}p_i A_i \left(\frac{(\alpha_i p_i A_i)}{\left( \sum_{i=1}^{S}(\alpha_i p_iA_i)^{\frac{1}{\alpha_1+1}}\right)^{\alpha_1+1}}\right)^{-\frac{\alpha_i}{\alpha_i+1}} \right).
    \end{align*}
    Therefore, taking $N$ large enough so that $\tilde{L}(\bar{\vq})<\min_i\{1,\frac{A_i}{B_i} \}$ shows that $\tilde{L}$ at $\bar{\vq}$ is smaller than $\tilde{L}$ for any $\vq$ with one of $\vq_i=0$. For this it suffices to take 

    \begin{align}\label{eq:second-N-condition}
        N>2 \left(\frac{1}{\min_i\{1,\frac{A_i}{B_i}\} } \left(\sum_{i=1}^{K}p_i A_i \left(\frac{(\alpha_i p_i A_i)}{\left( \sum_{i=1}^{S}(\alpha_i p_iA_i)^{\frac{1}{\alpha_1+1}}\right)^{\alpha_1+1}}\right)^{-\frac{\alpha_i}{\alpha_i+1}} \right)^{-1}\right)^{\frac{1}{\alpha_1}}.
    \end{align}
    Therefore, we have shown that for $N$ larger than the expressions in \Cref{eq:first-N-cond} and \Cref{eq:second-N-condition}, $\tilde{\vq}^*$ has no zero coordinates.

    Now we can find $\tilde{\vq}^*$ inside $(0,1)^K$ using Lagrange multipliers. Note that each $f_i$ is continuously differentiable on $(0,1)^K$, which is an open set containing the feasible set. Note that the constraint now is $\sum_{i=1}^{K}q_i-1=0$. We have that there exists $\lambda>0$ such that 
    \begin{align*}
        \frac{p_iA_i}{((Nq_i)^{\alpha_i}+B_i)^2}\alpha_i N^{\alpha_i} q_i^{\alpha_i-1}&=\lambda \\
        \sum_{i=1}^{K} q_i = 1.
    \end{align*}
    Note that this equation has a unique solution in $(0,\infty)$ for fixed $\lambda$ since $\frac{A_i}{((Nq_i)^{\alpha_i}+B_i)^2}\alpha_i N^{\alpha_i}$ is a decreasing function and $\lambda q_i^{1-\alpha_i}$ is an increasing function, and for $q_i = 0$ we have that $\frac{A_i}{B_i}^2 \alpha_i N^{\alpha_i}>0$. Let $\lambda^*$ and $\tilde{q}_i = \tilde{q}_i(\lambda^*)$ be the unique solution. 
    Note that $\tilde{L}(\tilde{q}^*)\le \tilde{L}(\bar{q})$ so in particular we have that 
    \begin{align*}
        p_i \frac{A_i}{(Nq_i)^{\alpha_i}+B_i}\le C \frac{1}{N^{\alpha_1}},
    \end{align*}
    where $C=2^{\alpha_1}\left(\sum_{i=1}^{K}p_i A_i \left(\frac{(\alpha_i p_i A_i)}{\left( \sum_{i=1}^{S}(\alpha_i p_iA_i)^{\frac{1}{\alpha_1+1}}\right)^{\alpha_1+1}}\right)^{-\frac{\alpha_i}{\alpha_i+1}} \right)$. So we have that 
    \begin{align*}
        \frac{p_i A_i }{C} N^{\alpha_1} &\le (Nq_i)^{\alpha_i}+B_i\\
        Nq_i &\ge \left( \frac{p_i A_i }{C} N^{\alpha_1}-B_i\right)^{\frac{1}{\alpha_i}}.
    \end{align*}
    Taking \begin{align*}
        N \ge \frac{2CB_i}{p_iA_i}
    \end{align*}
    for all $i$.
    Therefore, we have that 
    \begin{align*}
        Nq_i &\ge \left( \frac{1}{2}\frac{p_i A_i }{C} N^{\alpha_1}\right)^{\frac{1}{\alpha_i}}\\
        q_i &\ge N^{\frac{\alpha_1-\alpha_i}{\alpha_i}}
    \end{align*}
    Therefore as long as 
    \begin{align}\label{eq:third-N-condition}
        N>\left(\max_i\{\frac{2B_iC}{p_iA_i} \}\right)^{\frac{1}{\alpha_1}}
    \end{align}
    we have that 
    \begin{align*}
        \frac{A_i}{((Nq_i)^{\alpha_i}+B_i)^2} &\ge \frac{p_iA_i}{(Nq_i)^{\alpha_i}}(1-\frac{B_i}{(Nq_i)^{\alpha_i}})^2\ge \frac{p_iA_i}{(Nq_i)^{\alpha_i}} (1-\frac{2B_iC}{p_iA_iN^{\alpha_1}})^2\\
        &\ge \frac{p_iA_i}{(Nq_i)^{\alpha_i}} (1-\frac{4B_iC}{p_iA_iN^{\alpha_1}})
    \end{align*}
    for \begin{align}\label{eq:fourth-N-condition}
        N>\max_i \{ B_i^{\frac{1}{\alpha_i}}\}.
    \end{align}
    Therefore, the equation 
    \begin{align*}
        \frac{p_iA_i}{((Nq_i)^{\alpha_i}+B_i)^2}\alpha_i N^{\alpha_i} q_i^{\alpha_i-1}&=\lambda
    \end{align*}
    implies that
    \begin{align*}
        \frac{p_iA_i}{(Nq_i)^{2\alpha_i}}\alpha_i N^{\alpha_i} q_i^{\alpha_i-1}(1-\frac{4B_iC}{p_iA_iN^{\alpha_1}}) \le \lambda \le \frac{p_iA_i}{(Nq_i)^{2\alpha_i}}\alpha_i N^{\alpha_i} q_i^{\alpha_i-1}.
    \end{align*}
    Therefore, for all $q$ we have that 
    \begin{align*}        \left(\frac{p_iA_i\alpha_i}{N^{\alpha_i}\lambda}\right)^{\frac{1}{\alpha_i+1}}\left(1-\frac{4B_iC}{p_iA_iN^{\alpha_1}}\right)^{\frac{1}{\alpha_i+1}}\le q_i \le \left(\frac{p_iA_i\alpha_i}{N^{\alpha_i}\lambda}\right)^{\frac{1}{\alpha_i+1}}.
    \end{align*}
    Plugging this back into $\sum_{i=1}^{K} q_i = 1$ we have that for $\lambda$ it holds that 
    \begin{align*}
\sum_{i=1}^{K}\left(\frac{p_iA_i\alpha_i}{N^{\alpha_i}\lambda}\right)^{\frac{1}{\alpha_i+1}}\left(1-\frac{4B_iC}{p_iA_iN^{\alpha_1}}\right)^{\frac{1}{\alpha_i+1}}\le 1 \le\sum_{i=1}^{K} \left(\frac{p_iA_i\alpha_i}{N^{\alpha_i}\lambda}\right)^{\frac{1}{\alpha_i+1}}
    \end{align*}
    Therefore, we have that 
    \begin{align*}
        \lambda^* = \frac{1}{N^{\alpha_1}}\left(\sum_{i=1}^{S}(\alpha_ip_iA_i)^{\frac{1}{\alpha_i+1}}\right)^{\alpha_1+1}+O(\frac{1}{N^{2\alpha_1}}).
    \end{align*}
    From this we can compute that 
    \begin{align}\label{eq:final-tilde-q}
        \tilde{q}_i^*=\frac{1}{N^{\frac{\alpha_i-\alpha_1}{\alpha_i+1}}}\left(\frac{(\alpha_i p_i A_i)}{\left( \sum_{i=1}^{S}(\alpha_i p_iA_i)^{\frac{1}{\alpha_1+1}}\right)^{\alpha_1+1}}\right)^{\frac{1}{\alpha_i+1}}+O\left(\frac{1}{N^{\frac{\alpha_i-\alpha_1+2\alpha_1}{\alpha_i+1}}}\right). 
    \end{align}
    This finishes the proof. The lower bound on $N$  i.e. $N_0(p_i,A_i,B_i,\alpha_i)$ is given by the minimum of 
    \Cref{eq:first-N-cond,eq:second-N-condition,eq:third-N-condition} and \Cref{eq:fourth-N-condition}.
    This shows that 
    \begin{align*}
        \tilde{L}(\tilde{\vq}^*)  &= \frac{1}{N^{\alpha_1}} \left(\sum_{i=1}^{S} (\alpha_i p_i A_i)^{\frac{1}{\alpha_i+1}} \right)^{\alpha_1} \left(\sum_{i=1}^{S} \frac{(p_iA_i)^{\frac{1}{\alpha_i+1}}}{\alpha_i^{\frac{\alpha_i}{\alpha_i+1}}} \right)+ O\left(  \frac{1}{N^{\alpha_1+\frac{2\alpha_1^2}{\alpha_1+1}}}\right)
    \end{align*}
\end{proof}

\begin{proposition}[Approximate Optimal is Close to Optimal]\label{lemma:app:optimumsareclose}
    Let $\tilde{\vq}^*$ be the minimum of the approximate population error $\tilde{L}(\vq)$ in \Cref{eq:approximate-loss} and let $\vq^*$ be the minimum of the loss in Power Law \Cref{model:generalpowerlaw}. Then if $N\ge N_1(p_i,A_i,B_i,\alpha_i)$
    \begin{align*}
        L(\vq^*)&= \frac{1}{N^{\alpha_1}} \left(\sum_{i=1}^{S} (\alpha_i p_i A_i)^{\frac{1}{\alpha_i+1}} \right)^{\alpha_1} \left(\sum_{i=1}^{S} \frac{(p_iA_i)^{\frac{1}{\alpha_i+1}}}{\alpha_i^{\frac{\alpha_i}{\alpha_i+1}}} \right)+ O\left(  \frac{1}{N^{\alpha_1+\frac{2\alpha_1^2}{\alpha_1+1}}}\right)\\
        |\tilde{q}_i^*-q_i^*|&\le o(\frac{1}{N^{\frac{\alpha_i-\alpha_1}{\alpha_i+1}}})\\
        q_i^* &=\frac{1}{N^{\frac{\alpha_i-\alpha_1}{\alpha_i+1}}}\left(\frac{(\alpha_i p_i A_i)}{\left( \sum_{i=1}^{S}(\alpha_i p_iA_i)^{\frac{1}{\alpha_1+1}}\right)^{\alpha_1+1}}\right)^{\frac{1}{\alpha_i+1}}+o\left(\frac{1}{N^{\frac{\alpha_i-\alpha_1}{\alpha_i+1}}}\right).
    \end{align*}
\end{proposition}

\begin{proof}[Proof of \Cref{lemma:app:optimumsareclose}]
    Note that by \Cref{lemma:app:sufficientfk}, we have that for $\tilde{\vq}^*$ defined in \Cref{eq:final-tilde-q} 
    \begin{align*}
        L(\tilde{\vq}^*) &\le \tilde{L}(\tilde{\vq}^*) + \sum_{k=1}^{K} p_k \left(320\min\{ \frac{A_k}{B_k},1\} \frac{1}{(N\tilde{q}_k)^2}+\frac{\alpha_k A_k }{(N\tilde{q}_k)^{\alpha_k+\frac{1}{4}}}  + \frac{\alpha_kA_k}{(N\tilde{q}_k)^{\alpha_k+\frac{1}{2}}} \right) \\
        &\le \tilde{L}(\tilde{q}^*)+\frac{C_L}{N^{\frac{\alpha_1+1}{\alpha_i+1}(\alpha_i+\frac{1}{4})}}\le \tilde{L}(\tilde{q}^*)+\frac{C_L}{N^{\alpha_1+\frac{1}{4}}},
    \end{align*}
    where $C_L = 320\min\{ \frac{A_k}{B_k},1\}+2\alpha_k A_k$. Note additionally that by analogous logic from \Cref{lemma:app:sufficientfk} the inequality also holds the other way. By \Cref{lemma:app:sufficientfk}, we have that 
    \begin{align*}
        L(\vq^*)\ge \tilde{L}(\vq^*) - \sum_{k=1}^{K} p_k \left(320\min\{ \frac{A_k}{B_k},1\} \frac{1}{(N\tilde{q}_k)^2}+\frac{\alpha_k A_k }{(N\tilde{q}_k)^{\alpha_k+\frac{1}{4}}}  + \frac{\alpha_kA_k}{(N\tilde{q}_k)^{\alpha_k+\frac{1}{2}}} \right).
    \end{align*}
    Note that since $\tilde{L}(\tilde{\vq}^*)$ is the minimum, we have that 
    \begin{align*}
        \tilde{L}(\vq^*)\ge \tilde{L}(\tilde{\vq}^*).
    \end{align*}
    Therefore, we conclude 
    \begin{align*}
        L(\vq^*) \ge \tilde{L}(\tilde{\vq}^*) -\sum_{k=1}^{K} p_k \left(320\min\{ \frac{A_k}{B_k},1\} \frac{1}{(N\tilde{q}_k)^2}+\frac{\alpha_k A_k }{(N\tilde{q}_k)^{\alpha_k+\frac{1}{4}}}  + \frac{\alpha_kA_k}{(N\tilde{q}_k)^{\alpha_k+\frac{1}{2}}} \right).
    \end{align*}
    This fnishes the proof of the first claim that 
    \begin{align*}
        L(\vq^*)&= \frac{1}{N^{\alpha_1}} \left(\sum_{i=1}^{S} (\alpha_i p_i A_i)^{\frac{1}{\alpha_i+1}} \right)^{\alpha_1} \left(\sum_{i=1}^{S} \frac{(p_iA_i)^{\frac{1}{\alpha_i+1}}}{\alpha_i^{\frac{\alpha_i}{\alpha_i+1}}} \right)+ O\left(  \frac{1}{N^{\alpha_1+\frac{2\alpha_1^2}{\alpha_1+1}}}\right). \\ 
    \end{align*}
    Note that the above equations imply that 
    \begin{align*}
        |\tilde{L}(\tilde{\vq}^*)-\tilde{L}(\vq^*)| \le 2 \sum_{k=1}^{K} p_k \left(320\min\{ \frac{A_k}{B_k},1\} \frac{1}{(N\tilde{q}_k)^2}+\frac{\alpha_k A_k }{(N\tilde{q}_k)^{\alpha_k+\frac{1}{4}}}  + \frac{\alpha_kA_k}{(N\tilde{q}_k)^{\alpha_k+\frac{1}{2}}} \right) \le 2\frac{C_L}{N^{\alpha_1+\frac{1}{4}}}.
    \end{align*}
    Note now that for all $k$ we have that for all $q,q+h \in (0,1)$ that there is $\xi \in (q+h,q)$ with
    \begin{align*}
        f_k(q+h)-f_k(q) = f_k'(\xi) h.
    \end{align*}
    Therefore, for $k=1,\dots,S$ we have that 
    \begin{align*}
        f_i(\tilde{q}_i^*+h)-f_i(\tilde{q}_i^*) = f_i'(\xi_i) h
    \end{align*}
    for some $\xi_i \in (\tilde{q}_i^*,\tilde{q}_i^*+h)$.
    Therefore, 
    \begin{align*}
        |f_i(q_i^*)-f_i(\tilde{q}_i^*)| = |f_i'(\xi_i)| |q_i^*-\tilde{q}_i^*|.
    \end{align*}
    If for $i=1,2,\dots, S$ we have that $q_i^*>2\tilde{q}_i^*$, say $i=1$, there there exists index $j$ such that $q_j^*<\tilde{q}_i^*-\frac{\tilde{q}_1^*}{K}$. Note that all $|f_i'(x)|$ are decreasing, so then $|f_j'(\xi_j)|\ge |f_j'(\tilde{q}_j^*)|=|\lambda|\ge \frac{1}{N^{\alpha_1+\frac{1}{8}}}$ for $N$ large enough, i.e. it suffices to have 
    \begin{align}\label{eq:fifth-N-cond}
        N>16\left(\sum_{i=1}^{S}(\alpha_ip_iA_i)^{\frac{1}{\alpha_i+1}}\right)^{-8(\alpha_1+1)}.
    \end{align}
    Then we have that 
    \begin{align*}
         \frac{p_iC_0}{N^{\alpha_1+\frac{1}{8}}}\le \frac{p_i}{N^{\alpha_1+\frac{1}{8}}}|\tilde{q}_i^*|\le p_i |f_i(q_i^*)-f_i(\tilde{q}_i^*)|\le |\tilde{L}(\tilde{\vq}^*)-\tilde{L}(\vq^*)|\le \frac{2C_L}{N^{\alpha_1+\frac{1}{4}}},
    \end{align*}
    which is impossible for 
    \begin{align}\label{eq:sixth-N-cond}
        N>\left(\max_i\{ \frac{2C_L}{p_i}\}\right)^{8}.
    \end{align}
    Therefore, for all $i=1,\dots,S$ we have that $q_i^* \le 2\tilde{q}_i^*$. Therefore, we have that for all $i=1,\dots,S$, $|f_i'(\xi)|\ge \frac{1}{2^{2\alpha_1}} |f_i'(\tilde{q}_i^*)| =\frac{1}{2^{2\alpha_1}} |\lambda| \ge \frac{1}{2^{2\alpha_1}} \frac{1}{N^{\alpha_1}} C_{\lambda} $, where $C_{\lambda}=\left(\sum_{i=1}^{S}(\alpha_ip_iA_i)^{\frac{1}{\alpha_i+1}}\right)^{(\alpha_1+1)}$. Therefore, for all $i=1,\dots, S$ we have that 
    \begin{align*}
       \frac{1}{2^{2\alpha_1}} \frac{1}{N^{\alpha_1}} C_{\lambda} |q_i^*-\tilde{q}_i^*|\le p_i|f_i(q_i^*)-f_i(\tilde{q}_i^*)| |\tilde{L}(\tilde{\vq}^*)-\tilde{L}(\vq^*)|\le 2\frac{C_L}{N^{\alpha_1+\frac{1}{4}}}.
    \end{align*}
    Therefore, for all $i=1,\dots,S$ we have that 
    \begin{align*}
        |q_i^*-\tilde{q}_i^*|<\frac{2^{2\alpha_1+1}C_L}{C_{\lambda}N^{\frac{1}{4}}}.
    \end{align*}
    This shows that for $i=1,\dots,S$
    \begin{align*}
        q_i^* &=\frac{1}{N^{\frac{\alpha_i-\alpha_1}{\alpha_i+1}}}\left(\frac{(\alpha_i p_i A_i)}{\left( \sum_{i=1}^{S}(\alpha_i p_iA_i)^{\frac{1}{\alpha_1+1}}\right)^{\alpha_1+1}}\right)^{\frac{1}{\alpha_i+1}}+o\left(\frac{1}{N^{\frac{\alpha_i-\alpha_1}{\alpha_i+1}}}\right).
    \end{align*}
\end{proof}

\begin{proof}[Proof of \Cref{thrm:main:generalpowerlaw}]
Follows directly from \Cref{lemma:app:optimumsareclose} and \Cref{lemma:app:soltoapprox}.

\begin{proposition}[Minimizer is in the interior]\label{lemma:interiorminimizer}
Consider the approximate population error given by \Cref{eq:approximate-loss}. Let $\vq^*$ be the minimum on $\Delta^{K-1}$. Then it holds that $\vq^*_{i}\neq 0$ for all $i$ for which $\alpha_i<1$.
\end{proposition}
\begin{proof}[Proof of \Cref{lemma:interiorminimizer}]
    Assume that $\vq^*$ is such that $q_i^*=0$ with $\alpha_i<1$ for $i\in I$, where $I$ is a set of indices. There exists $j$ with $q_j^*\neq 0$ since $\sum_{i=1}^{K}q_i^*=1$. Consider the following function 
    \begin{align*}
        g(x) = \sum_{i\in I} \frac{p_iA_i}{(xN)^{\alpha_i}+B_i} + \frac{p_jA_j}{((q_j^*-|I|x)N)^{\alpha_j}+B_j}.
    \end{align*}
    Note that 
    \begin{align*}
        g'(x) = -\sum_{i \in I}\frac{p_iA_i}{\left((xN)^{\alpha_i}+B_i)\right)^{2}}\alpha_iN^{\alpha_i} x^{\alpha_i-1}  + \frac{p_jA_j}{\left( ((q_j^*-|I|x)N)^{\alpha_j}+B_j\right)^2}\alpha_j N^{\alpha_j}(q_j^*-|I|x)^{\alpha_j-1}.
    \end{align*}
    for $x\in (0,\frac{q_j^*}{|I|})$. Note also that on $x\in (0,\frac{q_j^*}{|I|})$, the function $g(x)$ is continous. We have that $\lim_{x\to 0+}g'(x) = -\infty$. There exists $0<\delta<\frac{q_j^*}{|I|}$ such that $g'(x)<0$ for all $x\in (0,\delta)$. To see this, not that if this were not the case, there would have to exist a sequence of points $x_1,x_2,\dots$ such that $x_i=0$.To see why, note that $\lim_{x\to 0}g'(x)=-\infty$ implies that if $g'(x_0)>0$ then there is $\tilde{x}_0 \in (0,x_0)$ with $g'(\tilde{x}_0)<0$ and so by IVT we have that there has to exist $x_1 \in (\tilde{x}_0,x_0)$ with $g'(x_1)=0$. Repeated this procedure gives the sequence $x_1,x_2,\dots$. This is a contradiction since $\lim_{n\to\infty} g'(x_n)=0$. Therefore, we have that $g(x)$ is decreasing on $(0,\delta)$. Assume that $g(0) \le g(x)$ for all $x\in (0,\delta)$. Therefore, we have that $\frac{g(x)-g(0)}{x}\ge 0$ for all $x \in (0,\delta)$. By MVT, for each $x\in (0,\delta)$ there exists $\xi_x \in (0,x)$ with $g'(\xi_x) = \frac{g(x)-g(0)}{x} \ge 0$. Again, this is a contradiction, since for $x\to 0+$ we have that $\xi_x\to 0+$ so in particular $0\le \lim_{x\to 0+} g'(\xi_x) = \lim_{x\to 0+} g'(x) = -\infty$. Therefore, there exists $y \in (0,\delta)$ with $g(0)>g(y)$. This contradicts the assumption that $q_i^*=0$ for all $i\in I$ because if $\tilde{\vq}_i = \begin{cases} q_i^* & i\neq j, i\notin I \\ y & i\in I \\ q_j^*-|I|y & i=j \end{cases}$, then $\tilde{L}(\tilde{\vq},\vp)<\tilde{L}(\vq^*,\vp)$. Therefore, $\vq^*$ has nonzero coordinates for all $q_i^*$ for which $\alpha_i \neq 1$.
\end{proof}

\end{proof}

\begin{proof}[Proof of \Cref{cor:main:generalpower}]
From \Cref{thrm:main:generalpowerlaw}, by directly plugging in we have that since here $S=K$
    \begin{align*}
        q^*_{i} &= \frac{p_i^{\frac{1}{\alpha+1}}}{\sum_{i=1}^{m} p_i^{\frac{1}{\alpha+1}}}+o(1)\\
        q^*_1 &= \frac{p^{\frac{1}{\alpha+1}}}{p^{\frac{1}{\alpha+1}}+(K-1)\left(\frac{1-p}{K-1} \right)^{\frac{1}{\alpha+1}}}+o(1)\\
        q_{i\ge 2}&= \frac{\left(\frac{1-p}{K-1} \right)^{\frac{1}{\alpha+1}}}{p^{\frac{1}{\alpha+1}}+(K-1)\left(\frac{1-p}{K-1} \right)^{\frac{1}{\alpha+1}}}+o(1)
        \end{align*}
        Therefore, this immediately shows the claim about $q_i^*$.
        Therefore, we have that 
        \begin{align*}
            N^{\mathrm{ratio}}
=
\left(
  \frac{
    \bigl(p^{\frac{1}{\alpha+1}} + (K-1)\bigl(\tfrac{1-p}{K-1}\bigr)^{\frac{1}{\alpha+1}}\bigr)^{\alpha+1}
  }{
    p^{1-\alpha} + (K-1)^{\alpha}(1-p)^{1-\alpha}
  }
\right)^{\frac{1}{\alpha}}+o(1).
        \end{align*}
        The only thing left to prove is the inequality. Let $\delta = \left(\frac{p}{1-p}\right)^{\frac{1}{\alpha+1}}(K-1)^{-\frac{\alpha}{\alpha+1}}$. Note that we can write 
        \begin{align*}
            p^{\frac{1}{\alpha+1}} + (K-1)\bigl(\tfrac{1-p}{K-1}\bigr)^{\frac{1}{\alpha+1}} = (K-1)^{\frac{\alpha}{\alpha+1}}(1-p)^{\frac{1}{\alpha+1}}\left(1+\delta\right).
        \end{align*}
        Note that $p^{1-\alpha} + (K-1)^{\alpha}(1-p)^{1-\alpha}\ge (K-1)^{\alpha}(1-p)^{1-\alpha}$.
        Therefore, we can write that 
        \begin{align*}
            N^{\textrm{ratio}} \le (1-p)(1+\delta)^{\frac{\alpha+1}{\alpha}}+o(1).
        \end{align*}
        Note also that $(1+\delta)^{t}\le 1+(2^t-1)\delta$ for $\delta<1$, since $f(x)=(1+x)^t$ has $f''(x)=t(t-1)(1+x)^{t-2}$ so for $t>1$ it is convex. Therefore, $f(\delta)\le f(0)+(f(1)-f(0))\delta = 1+(2^t-1)\delta$. Using this for $t=\frac{\alpha+1}{\alpha}$, we have that  
        \begin{align*}
        N^{\textrm{ratio}}(\vp) & \le (1-p)+(2^{\frac{\alpha+1}{\alpha}}-1)\left(\frac{p}{1-p}\right)^{\frac{1}{\alpha+1}}K^{-\frac{\alpha}{\alpha+1}}+o(1).
    \end{align*}
    So it suffices to have the $o(1)$ term be smaller than $\left(\frac{p}{1-p}\right)^{\frac{1}{\alpha+1}}K^{-\frac{\alpha}{\alpha+1}}$. Note that from the proof of \Cref{lemma:app:optimumsareclose}, we can compute the $o(1)$ term. In $q_i^*$, the term was bounded by $\frac{2^{2\alpha_1+1}C_L}{C_{\lambda}N^{\frac{1}{4}}}$ where $C_L$ and $C_{\lambda}$ can be written explpicitly in terms of $A,B,\alpha,K,p$. In $N^{\textrm{ratio}}$ the consants are additionally mutliplied by $\frac{1}{\alpha}\left(
  \frac{
    \bigl(p^{\frac{1}{\alpha+1}} + (K-1)\bigl(\tfrac{1-p}{K-1}\bigr)^{\frac{1}{\alpha+1}}\bigr)^{\alpha+1}
  }{
    p^{1-\alpha} + (K-1)^{\alpha}(1-p)^{1-\alpha}
  }
\right)$, so it suffices to have 
\begin{align*}
    \frac{1}{\alpha}\left(
  \frac{
    \bigl(p^{\frac{1}{\alpha+1}} + (K-1)\bigl(\tfrac{1-p}{K-1}\bigr)^{\frac{1}{\alpha+1}}\bigr)^{\alpha+1}
  }{
    p^{1-\alpha} + (K-1)^{\alpha}(1-p)^{1-\alpha}
  }
\right)\frac{2^{2\alpha_1+1}C_L}{C_{\lambda}N^{\frac{1}{4}}} &\le  \left(\frac{p}{1-p}\right)^{\frac{1}{\alpha+1}}K^{-\frac{\alpha}{\alpha+1}}\\
\frac{1}{\alpha}\left(
  \frac{
    \bigl(p^{\frac{1}{\alpha+1}} + (K-1)\bigl(\tfrac{1-p}{K-1}\bigr)^{\frac{1}{\alpha+1}}\bigr)^{\alpha+1}
  }{
    p^{1-\alpha} + (K-1)^{\alpha}(1-p)^{1-\alpha}
  }
\right)\frac{2^{2\alpha_1+1}C_L}{C_{\lambda}N^{\frac{1}{4}}}\left(\frac{p}{1-p}\right)^{-\frac{1}{\alpha+1}}K^{\frac{\alpha}{\alpha+1}} &\le N^{\frac{1}{4}}.
\end{align*}
This happens when $N>N_0(p,\alpha,A,K,B)$.
    This finishes the proof.
\end{proof}

\subsection{Memorization Tasks}\label{app:orthogonalmemorization}

\begin{proof}[Proof of \Cref{thrm:main:orthogonalmemorization}]
Follows from \Cref{lm:qkopt-main} and \Cref{lemma:app:orthogonalmem-errors}.
\end{proof}

\begin{proof}[Proof \Cref{cor:main:orthogonalmemorization}]

First, note that in this case $\left(\frac{p_K}{p_k}\right)^{\frac{1}{N-1}}=\Theta((\frac{k}{K})^{\alpha/(N-1)})$. Therefore, only for $l=\Theta(K)$ do we have $f_N(l) = \Theta(1)$, so indeed then $K_N = \Theta(K)$ in this case. We directly compute that $\Lsame(\vp) = \Theta(N^{-1+\frac{1}{\alpha}})$. $L^*(\vp)$ follows directly from \Cref{lemma:app:orthogonalmem-errors} by using $K_N=\Theta(K)$, and we get $L^*(\vp) = \Theta(N^{\alpha-1})$

\end{proof}

\paragraph{Proofs for the Memorization Case }


For every task $k$, we only need to memorize the unique hypothesis that appears together with the task.
\begin{align*}
    \bar{e}_k(\vq) =(1 - q_k)^N, &\qquad L_N(\vp,\vq)=  \sum_{k=1}^{K} p_k (1-q_k)^N.
\end{align*}
Let $\{q^*_k(N)\}_{k=1}^{M} = \argmin_{\{q_k\}_{k=1}^{M}} \left\{ L_N(\vp,\vq) \right\}$.

\begin{lemma}\label{lm:qkopt-main}
    For all $N \ge 1$, there exists $\beta_N > 0$ such that the following holds for $q^*_k(N)$:
    \begin{align*}
        q^*_k(N) = \max\left\{0, 1 - \beta_N \cdot p_k^{-1/(N-1)}\right\}.
    \end{align*}
\end{lemma}
\begin{proof}
    By the method of Lagrange multipliers, there exists $\lambda \in \R$ such that
    \begin{align*}
        -N p_k (1 - q^*_k(N))^{N-1} + \lambda = 0, \qquad \forall k \in [M] \quad \text{s.t.} \quad q^*_k(N) > 0.
    \end{align*}
    Then we have
    \begin{align*}
        q^*_k(N) = 1 - \left(\frac{\lambda}{N p_k}\right)^{1/(N-1)}.
    \end{align*}
    Setting $Z_N := \left(\frac{N}{\lambda}\right)^{1/(N-1)}$ and $\beta_N=\frac{1}{Z_N}$ finishes the proof.
\end{proof}

Let $K_N := \max\{ k \in [K] : q_k^*(N) \ne 0\}$. $K_N$ and $\beta_N$ satisfy the following relationship.
\begin{lemma}
    For all $N \ge 1$,
\begin{align*}
    \beta_N &= \frac{K_N - 1}{\sum_{k=1}^{K_N} p_k^{-1/(N-1)}} \in \left[p_{K_N+1}^{1/(N-1)}, p_{K_N}^{1/(N-1)}\right), \\
    K_N &= \max\{l \mid f_N(l) < 1\} \quad \text{where} \quad f_N(l) := \sum_{k=1}^{l-1} \left(
        1 - \left( \frac{p_{K}}{p_{k}} \right)^{1/(N-1)}
    \right).
\end{align*}
\end{lemma}
\begin{proof}
    Since $\sum_{k=1}^{K} q^*_k(N) = 1$ and $q_k^*(N) = 0$ for all $k > K_N$, by~\Cref{lm:qkopt-main}, we have
    \begin{align*}
        \sum_{k=1}^{K_N} \left( 1 - \beta_N \cdot p_k^{-1/(N-1)} \right) = 1.
    \end{align*}
    Rearranging the terms, we obtain
    \begin{align*}
        \beta_N \sum_{k=1}^{K_N} p_k^{-1/(N-1)}  = K_N - 1,
    \end{align*}
    which implies $\beta_N = \frac{K_N - 1}{\sum_{k=1}^{K_N} p_k^{-1/(N-1)}}$.
    
    By definition of $K_N$, $1 - \beta_N \cdot p_{K_N}^{-1/(N-1)} > 0$ and $1 - \beta_N \cdot p_{K_N+1}^{-1/(N-1)} \le 0$. This implies $\beta_N \in [p_{K_N+1}^{1/(N-1)}, p_{K_N}^{1/(N-1)})$. Then we have
    \begin{align*}
        1 &> \sum_{k=1}^{K_N} \left( 1 - p_{K_N}^{1/(N-1)} \cdot p_k^{-1/(N-1)} \right) = \sum_{k=1}^{K_N-1} \left( 1 - p_{K_N}^{1/(N-1)} \cdot p_k^{-1/(N-1)} \right). \\
        1 &\le \sum_{k=1}^{K_N} \left( 1 - p_{K_N+1}^{1/(N-1)} \cdot p_k^{-1/(N-1)} \right).
    \end{align*}
    Let $f_N(K) := \sum_{k=1}^{K-1} \left(
        1 - p_{K}^{1/(N-1)} \cdot p_k^{-1/(N-1)}
    \right)$. Then $K_N = \max\{K : f_N(K) < 1\}$.
\end{proof}
\begin{lemma}\label{lemma:app:orthogonalmem-errors}
    Test errors for sampling with $\vq = \vp$ and $\vq = \vq^*$ are
    \begin{align*}
        \Lsame(\vp) &= \sum_{k=1}^{K} p_k (1-p_k)^N, \\
        L^*(\vp) &= \sum_{k=K_N+1}^{K} p_k+(K_N - 1)\beta_N^{N-1} \in \left[
            \sum_{k=K_N+1}^{K} p_k + (K_N - 1) p_{K_N + 1},
            \sum_{k=K_N+1}^{K} p_k + (K_N - 1) p_{K_N}
        \right).
    \end{align*}
\end{lemma}
\begin{proof}
    The first equation is straightforward. For the second equation,
    \begin{align*}
        L^*(\vp) = \sum_{k=K_N+1}^{K} p_k+\sum_{k=1}^{N} p_k (1-q^*_k)^N &= \sum_{k=1}^{K_N} p_k (\beta_N \cdot p_k^{-1/(N-1)})^N \\
        &= \sum_{k=K_N+1}^{K} p_k+\beta_N^N\sum_{k=1}^{K_N} p_k^{-1/(N-1)}\\
        &= \sum_{k=K_N+1}^{K} p_k+ \beta_N^N \cdot \left( \frac{K_N-1}{\beta_N}\right) \\
        &= \sum_{k=K_N+1}^{K} p_k+(K_N - 1)\beta_N^{N-1}.
    \end{align*}
    Further noting that $\beta_N \in \left[p_{K_N+1}^{1/(N-1)}, p_{K_N}^{1/(N-1)}\right)$ completes the proof.
\end{proof}

\section{Proof of the Existence of PDS in the General Case}\label{app:generalcase}

\subsection{Proof of Main Theorem}

We provide a functional-analytic characterization of when positive distribution shift is guaranteed to exist. The key idea is to study the loss $\Lshift(\vp, \vr)$ as a function of both the target mixing ratios $\vp$ and the training mixing ratios $\vr$, and show that $\vr = \vp$ almost never minimizes $\Lshift(\vp, \vr)$ except for the degenerate cases described in~\Cref{thm:main}.

The following key property of $f_k$ is useful for our analysis:
\begin{lemma}\label{lm:f_k_property}
    For all $k \in [m]$, the function $f_k$ is a $0$-homogeneous rational function.
\end{lemma}
\begin{proof}
    It is easy to see that $f_k$ is $0$-homogeneous, since by definition,
    it holds for all $c > 0$ that
    $f_k(c\vr) = f_k(\frac{c\vr}{\abs{c \vr}}) = f_k(\vr)$.

    To show that $f_k$ is a rational function, recall that for $\vr \in \Delta^{m-1}$, $f_k(\vr)$ is defined as the expected loss of the model trained on a dataset $S$ sampled with mixing ratio $\vr$ and evaluated on subpopulation $\gD_k$.

    Sampling from $\vr$ corresponds to first sampling subpopulation indices $i_1, \dots, i_n \in [m]$ according to $\vr$, and then drawing the $j$-th sample in the dataset $S$ from the subpopulation $\gD_{i_j}$.
    This allows us to rewrite the expectation as:
    \begin{align*}
        F_k(\vr) := \sum_{1 \le i_1, \dots, i_n \le m} \left( r_{i_1} \cdots r_{i_n} \cdot \E_{S \sim \gD_{i_1} \times \gD_{i_2} \times \cdots \times \gD_{i_n}}\E_{\vz \sim \gD_k}\left[  \ell(\gA(S), z)\right]\right)
    \end{align*}
    Each term in this sum $F_k(\vr)$ is the product of a monomial of degree $n$ in $\vr$ and a constant that does not depend on $\vr$.
    Therefore, $F_k(\vr)$ is a degree-$n$ polynomial in $\vr \in \Delta^{m-1}$.
    Since $f_k(\vr) := F_k( \frac{\vr}{\abs{\vr}})$ by definition, it follows that $f_k$ is a rational function on $\R^{m}_{\ge 0} \setminus \{\vzero\}$.
\end{proof}


Define the total population loss when testing under $\vp$ but training under $\vr$ as $\Lshift(\vp, \vr) := \sum_{k=1}^{m} p_k f_k(\vr)$.
We now characterize when $\vr = \vp$ is a minimizer of $\Lshift(\vp, \vr)$ over $\vr \in \Delta^{m-1}$.

\begin{lemma}\label{lm:gradient-condition}
For any $\vp \in \Delta_+^{m-1}$, if $\Lsame(\vp) = \Lshift^*(\vp)$,
then
\begin{equation}\label{eq:gradient-condition}
\sum_{k=1}^{m} p_k \frac{\partial f_k(\vp)}{\partial p_i}
=
0
\quad \text{for all } i \in [m].
\end{equation}
\end{lemma}

\begin{proof}
We minimize $\Lshift(\vp, \vr)$ over $\vr \in \Delta^{m-1}$ using the method of Lagrange multipliers. Define the Lagrangian:
$$
\cJ(\vr, \lambda) = \sum_{k=1}^{m} p_k f_k(\vr) - \lambda \left( \sum_{k=1}^{m} r_k - 1 \right).
$$
At a minimizer $\vr = \vp$, the stationarity condition requires $\frac{\partial}{\partial r_i}\cJ(\vr, \lambda) = 0$ for all $i \in [m]$. This yields
$$
\frac{\partial}{\partial r_i} \left( \sum_{k=1}^{m} p_k f_k(\vr) \right) \bigg|_{\vr = \vp} = \lambda \quad \text{for all } i \in [m].
$$
That is,
$$
\sum_{k=1}^{m} p_k \frac{\partial f_k(\vp)}{\partial p_i} = \lambda.
$$
Multiplying both sides by $p_i$ and summing over $i \in [m]$ gives:
\begin{align*}
    \sum_{i=1}^{m} p_i \lambda = \sum_{i=1}^{m} p_i \sum_{k=1}^{m} p_k \frac{\partial f_k(\vp)}{\partial p_i} 
    = \sum_{k=1}^{m} \bigg( p_k \cdot \inne{\vp}{\nabla f_k(\vp)}\bigg) 
    = \sum_{k=1}^{m} (p_k \cdot 0\big)
    = 0,
\end{align*}
where the third equality holds because $f_k$ is $0$-homogeneous and thus $\inne{\vp}{\nabla f_k(\vp)} = 0$ by Euler's theorem.
Thus, $\lambda = 0$, and we have $\sum_{k=1}^{m} p_k \frac{\partial f_k(\vp)}{\partial p_i} = 0$, as claimed.
\end{proof}

We now connect this condition to a gradient field characterization.
\begin{theorem}
\label{thm:gradient-equivalence}
For any learning algorithm $\gA$, one of the following two scenarios must hold:
\begin{enumerate}
\item $\Lsame(\vp) = \Lshift^*(\vp)$ holds only for a zero-measure subset of $\vp \in \Delta^{m-1}$;
\item $\nabla \Lsame(\vp) = (f_1(\vp), \dots, f_m(\vp))$.
\end{enumerate}
\end{theorem}

\begin{proof}
    Let $\Omega_{i}$ denote the set of $\vp \in \Delta_+^{m-1}$ for which the gradient condition~\eqref{eq:gradient-condition} holds for index $i \in [m]$.
    By~\Cref{lm:f_k_property}, the function $f_k$ is a rational function of $\vp$. It follows that both $\frac{\partial f_k(\vp)}{\partial p_i}$ and $\sum_{k=1}^{m} p_k \frac{\partial f_k(\vp)}{\partial p_i}$ are also rational functions of $\vp$. 
    Therefore, $\Omega_i$ is the zero set of a rational function, and must be either a measure-zero subset of $\Delta_+^{m-1}$ or the entire domain. 

    Let $\Omega := \bigcap_{i \in [m]} \Omega_{i}$ be the intersection of all $\Omega_{i}$. Then $\Omega$ is either a zero-measure subset of $\Delta_+^{m-1}$ or the entire domain.
    If $\Omega$ is a zero-measure subset, then by~\Cref{lm:gradient-condition}, we are in the first case of the theorem.
    If $\Omega$ is the entire domain, then the gradient condition~\eqref{eq:gradient-condition} holds for all $i \in [m]$, $\vp \in \R^m_{\ge 0} \setminus \{\vzero\}$.
    
    Recall that $\Lsame(\vp) := \sum_{k=1}^{m} p_k f_k(\vp)$.
    Then we compute:
    $$
    \frac{\partial \Lsame(\vp)}{\partial p_i} = f_i(\vp) + \sum_{k=1}^{m} p_k \frac{\partial f_k(\vp)}{\partial p_i}.
    $$
    By the gradient condition~\eqref{eq:gradient-condition}, $\sum_{k=1}^{m} p_k \frac{\partial f_k(\vp)}{\partial p_i} = 0$.
    Thus,
    $\frac{\partial \Lsame(\vp)}{\partial p_i} = f_i(\vp)$, which implies $\nabla \Lsame(\vp) = (f_1(\vp), \dots, f_m(\vp))$, which is the second case of the theorem.
\end{proof}
Finally, \Cref{thm:gradient-equivalence} implies \Cref{thm:main}.

\subsection{Characterization of Conservation Conditions}

\begin{proof}[Proof of \Cref{lemma:main:orthogonaltasks}]
    If~\Cref{cond:gradient} holds, then for all $i, j \in [m]$ ($i \ne j$),
    \begin{align*}
        \frac{\partial}{\partial p_j} f_i(\vp) = \frac{\partial^2}{\partial p_i \partial p_j} \Lsame(\vp) = \frac{\partial}{\partial p_i} f_j(\vp).
    \end{align*}
    By the chain rule, we have $\frac{\partial}{\partial p_j} f_i(\vp) = - \frac{p_i}{\abs{\vp}^2} g'_i(\frac{p_i}{\abs{\vp}})$ and $\frac{\partial}{\partial p_i} f_j(\vp) = - \frac{p_j}{\abs{\vp}^2} g'_j(\frac{p_j}{\abs{\vp}})$.
    Thus, for all $\vp \in \R^m_{\ge 0} \setminus \{\vzero\}$,
    \begin{align*}
        -\tfrac{p_i}{\abs{\vp}^2} g'_i(\tfrac{p_i}{\abs{\vp}}) = - \tfrac{p_j}{\abs{\vp}^2} g'_j(\tfrac{p_j}{\abs{\vp}}).
    \end{align*}
    For any $x, y > 0$ with $x + y < 1$, we can choose $\vp$ such that $\frac{p_i}{\abs{\vp}} = x$ and $\frac{p_j}{\abs{\vp}} = y$.
    Then we have $xg'_i(x) = yg'_j(y)$ for all such $x, y$.
    This is only possible if there exists a constant $C$ such that $xg'_i(x) = C$ for all $x \in (0, 1)$.
    Solving this gives $g'_i(x) = \frac{C}{x}$, which implies that $g_i(x) = C \ln x + A$ for some constant $A$.
    
    Since $g_i(x)$ has no singularity at $x = 0$, we must have $C = 0$.
    Thus, $g_i(x)$ is a constant function.
\end{proof}

Further, we show that if the Conservation \Cref{cond:gradient} is satisfied, then one function $f_i$ determines the rest up to a constant.

\begin{lemma}\label{lm:conservation-unique}
    If both $(f_1, \dots, f_K, \Lsame)$
    and $(\hat{f}_1, \dots, \hat{f}_K, \hLsame)$
    satisfy \Cref{cond:gradient},
    and if $f_i = \hat{f}_i$ for some $i \in [m]$,
    then for all $k \ne i$,
    $f_k(\vp) = \hat{f}_k(\vp) + C_k$ for some constant $C_k$.
\end{lemma}

The above \Cref{lm:conservation-unique} implies that for every $k$ and corresponding error function $e_k(\vn)$, there exists at most one tuple of error functions $\{e_j\}_{j=1,j\neq k}^K$ (up to a individual constant offset for each error function $e_j$) that positive distribution shift does not happen for $\vp$ of positive measure. This further implies the following corollary.

\begin{corollary}[Positive Distribution Shift \emph{Almost} Always Exists for General Tasks]\label{cor:main:almostalways} For any set of $K\ge 3$ subpopulations $\cD_1,\dots,\cD_K$ and any learning algorithm $\gA$, for all $\vp\in\Delta^{K-1}_+$, the configuration of $[e_k(\vn)]_{k\in[K],\vn}$ that positive distribution shift does not happen is zero-measure.
\end{corollary}

\Cref{cor:main:almostalways} shows that either the test mixing ratio $\vp$ is on a set of measure zero on the simplex or the configuration of subpopulation error functions $e_k(\vn)$ is on a set of measure zero. This implies that positive distribution shift exists \emph{almost} always.

\begin{proof}[Proof of \Cref{lm:conservation-unique}]
    Let $\Delta(\vp) = \Lsame(\vp) - \hLsame(\vp)$ be the difference between the two losses when training on the same distribution.
    By~\Cref{cond:gradient}, we have
    \begin{equation}\label{eq:conservation-unique-Lsame}
        \frac{\partial}{\partial p_i} \Delta(\vp) = \frac{\partial}{\partial p_i} \Lsame(\vp) - \frac{\partial}{\partial p_i} \hLsame(\vp)  = f_i(\vp) - \hat{f}_i(\vp) = 0.
    \end{equation}
    Therefore, $\Delta(\vp)$ is independent of $p_i$, and there exists a function $C: \R^{m-1} \to \R, \vp_{-i} \mapsto C(\vp_{-i})$ such that $\Delta(\vp) = C(\vp_{-i})$ for all $\vp \in \R^m_{\ge 0} \setminus \{\vzero\}$, where $\vp_{-i} = (p_1, \dots, p_{i-1}, p_{i+1}, \dots, p_m)$.
    This is because we can set $C(\vp_{-i}) = \left.\Delta(\vp)\right|_{p_i = 0}$
    and then take the integral of $\frac{\partial}{\partial p_i} \Delta(\vp)$ over $p_i$ to get $\Delta(\vp) = C(\vp_{-i})$.

    Next, note that both $\Lsame(\vp)$ and $\hLsame(\vp)$ can be written as 
    rational functions of the form
    \begin{align*}
        \Lsame(\vp) = \frac{S(\vp)}{\abs{\vp}^{n}}, \qquad \hLsame = \frac{\hat{S}(\vp)}{\abs{\vp}^{n}},
    \end{align*}
    where $n$ is the dataset size.
    This is because $\Lsame(\vp) = \sum_{k=1}^{m} p_k f_k(\vp)$.
    
    Now we show that $\Delta(\vp)$ must have the form $\Delta(\vp) = \sum_{k \ne i} C_k p_k$ for some constants $C_k$.
    Let $D(\vp) := S(\vp) - \hat{S}(\vp)$.
    Since $D(\vp)$ is a polynomial, $C(\vp_{-i}) = \frac{D(\vp)}{\abs{\vp}^n}$ must be a rational function. Let $C(\vp_{-i}) = \frac{A(\vp_{-i})}{B(\vp_{-i})}$ for some polynomials $A(\vp_{-i})$ and $B(\vp_{-i})$.
    Then
    \begin{align*}
        D(\vp) B(\vp_{-i}) = A(\vp_{-i}) \abs{\vp}^n.
    \end{align*}
    If $A = 0$, then $\Delta(\vp) = 0$.
    Otherwise, both $A(\vp_{-i})$ and $B(\vp_{-i})$ are non-zero polynomials.
    Since $B(\vp_{-i})$ cannot be divisible by $\abs{\vp}^n$, $D$ must be divisible by $\abs{\vp}^n$.
    Note that $D$ is a $(n+1)$-homogeneous polynomial
    and $\abs{\vp}^n$ is $n$-homogeneous,
    so $C(\vp_{-i}) = \frac{D}{\abs{\vp}^n}$ must be a $1$-homogeneous polynomial.
    The only $1$-homogeneous polynomials in variables $\vp_{-i}$ are linear functions of the form $C(\vp_{-i}) = \sum_{k \ne i} C_k p_k$ for some constants $C_k$.
    Thus, no matter $A = 0$ or not, 
    we have $\Delta(\vp) = \sum_{k \ne i} C_k p_k$.

    Finally, by~\Cref{cond:gradient}, we can compute for all $k \ne i$ that
    \begin{align*}
        f_k(\vp) - \hat{f}_k(\vp) =
        \frac{\partial}{\partial p_k} \Delta(\vp) = C_k,
    \end{align*}
    which implies that $f_k(\vp) = \hat{f}_k(\vp) + C_k$, as desired.
\end{proof}

\begin{proof}[Proof of \Cref{cor:main:almostalways}]
    This follows from \Cref{lm:conservation-unique}. Note that \Cref{lm:conservation-unique} implies that for every $k$ and corresponding error function $e_k(\vn)$, there exists at most one tuple of error functions $\{e_j\}_{j=1,j\neq k}^K$ (up to a individual constant offset for each error function $e_j$) that positive distribution shift does not happen for $\vp$ of positive measure. This implies the corollary.
\end{proof}

\section{Experiment Details}\label{app:experiment}

\paragraph{Model Architecture and Tokenizer.} We use a model architecture similar to GPT-2, except that we use RoPE instead of absolute position embedding. Our model has $6$ layers, $8$ attention heads, and $512$ embedding dimensions.
We use the same tokenizer as GPT-2, which is a byte-pair encoding (BPE) tokenizer.

\paragraph{Generation of Skills.} We randomly generate $M = 10^5$ skills. For each skill, we randomly sample $3$ English tokens and concatenate them to form the skill ID. The first token is sampled from a set of $1000$ tokens that start with a blank space and then a capital letter. The second and third tokens are sampled from a set of $1000$ tokens that start with a captial letter without a blank space.
The starting blank space is to ensure that the skill ID is tokenized into exactly $3$ tokens when placed in a prompt with space-separated skill IDs.
For example, ``\verb|CourtClientCheck|'' can be a skill ID (with blank space removed).
Then, for each skill $i$, we uniformly randomly sample a function $g_i$ that maps a number from $\{0, \dots, 9\}$ to $\{0, \dots, 9\}$.

\paragraph{Distribution: Skill Composition.} For each data point, a number $k$ is sampled uniformly from $\{10, \dots, 50\}$, then a set of $k$ skills $g_{i_1}, \dots, g_{i_k}$ are sampled IID following a power law $p(i) \propto (i + 50)^{-\alpha}$ with exponent $\alpha = 1.5$. The text consists of two parts. The input part is as follows:
\begin{center}
    \begin{verbatim}
    <|begin_of_text|> Input:
    [x] -> [skill ID 1] -> [skill ID 2] -> ... -> [skill ID k]\end{verbatim}
\end{center}
The output part is as follows:
\begin{center}
    \begin{verbatim}
    Output:
    [x] -> [skill ID 1] = [x1]
    [x1] -> [skill ID 2] = [x2]
    [x2] -> [skill ID 3] = [x3]
    ...
    [xk-1] -> [skill ID k] = [xk]
    [xk]\end{verbatim}
\end{center}
The input and output parts are concatenated together with a blank line in between.

\paragraph{Distribution: Uniform Skills.} For each data point, we randomly sample a skill ID uniformly from the skill ID set. Then the text is as follows:
\begin{center}
    \begin{verbatim}
    <|begin_of_text|> [x] [skill ID] = [expected output]\end{verbatim}
\end{center}

\paragraph{Evaluation.} We evaluate the test accuracy of the model on skill composition task with CoT reasoning. We sample $400$ data points from the skill composition task, but fix $k$ to be $10, 30, 50$. For each data point, only the input part is given to the model, and the model's autoregressive output is considered as correct if the last line of the output is the same as the expected output.

\paragraph{Training with Matched Distribution.} In the training, we use batch size $128$ and maximum sequence length $2048$. We perform sequence packing for each sequence in the batch: we sample data points from the skill composition task until the maximum sequence length is reached.
We train the model on $4$ A4000 GPUs for at most 40K steps.

\paragraph{Training with Mismatched Distribution.} Similar as above, but for every sequence in the batch, we first choose the task to be skill composition or uniform skills with probability $70\%$ and $30\%$ respectively. Then we sample data points from the chosen task until the maximum sequence length is reached. These sequences are then packed together to form the batch.

\paragraph{Results.} We show the results in \Cref{fig:skill-comp}. We see that training with mismatched distribution significantly outperforms training with matched distribution.

\newpage

\newpage

\end{document}